%% file: main.tex
\documentclass[10pt,twocolumn,letterpaper]{article}

\usepackage{cvpr}              
\usepackage{multirow}

\input{preamble}
\definecolor{cvprblue}{rgb}{0.21,0.49,0.74}
\usepackage[pagebackref,breaklinks,colorlinks,allcolors=cvprblue]{hyperref}
\usepackage[table]{xcolor}

\usepackage{etoc} 
\usepackage[table]{xcolor}
\usepackage{booktabs}
\usepackage{multirow}
\usepackage{graphicx}

\def\confName{CVPR}
\def\confYear{2027}

\title{Beyond Alignment: Supervising the Generator for EEG-to-Image Reconstruction}

\author{
Sangmin Jo \quad Wootaek Jeong \quad Heung-Il Suk\thanks{Corresponding author.}\\
Department of Artificial Intelligence, Korea University, Seoul, South Korea\\
{\tt\small \{sangminjo, wtjeong, hisuk\}@korea.ac.kr}
}

\etocdepthtag.toc{mainpart}
\begin{document}
\maketitle
\input{sec/0_abstract}    
\input{sec/1_intro}
\input{sec/2_rel}

\input{sec/3_preliminaries}

\input{sec/4_method}
\input{sec/5_experiment}

\input{sec/6_conclusion}

{   \small
    \bibliographystyle{ieeenat_fullname}
    \bibliography{main}
}


\input{sec/X_suppl}


\end{document}

%% file: sec/0_abstract.tex
\begin{abstract}
Visual brain decoding translates neural activity into human-interpretable images. While recent EEG-to-image reconstruction methods rely on aligning EEG signals with pretrained visual embeddings to condition an image generator, we empirically demonstrate that stronger cross-modal alignment does not translate into comparable gains in reconstruction quality. This indicates that improving the input condition alone is insufficient; the generative process itself requires explicit supervision. Although Representation Alignment (REPA) offers a solution by guiding diffusion transformers with pretrained vision encoder features, we find that EEG-conditioned generators struggle to follow these fixed targets. To address this mismatch, we introduce \textbf{G}en\textbf{E}r\textbf{A}tion-de\textbf{R}ived \textbf{S}upervision (GEARS): an image-conditioned teacher that translates static image-encoder features into the internal representation space of a diffusion transformer. Our EEG-conditioned student generator closely tracks these dynamic targets during denoising, discarding the teacher at inference. Evaluated on THINGS-EEG and THINGS-MEG, our approach outperforms conditioning-centric baselines, achieving state-of-the-art reconstruction fidelity. Notably, on THINGS-EEG, GEARS consistently improves reconstruction across diverse EEG representations.

\end{abstract}

%% file: sec/1_intro.tex
\begin{figure}[t]
    \centering
    \includegraphics[width=1\linewidth]{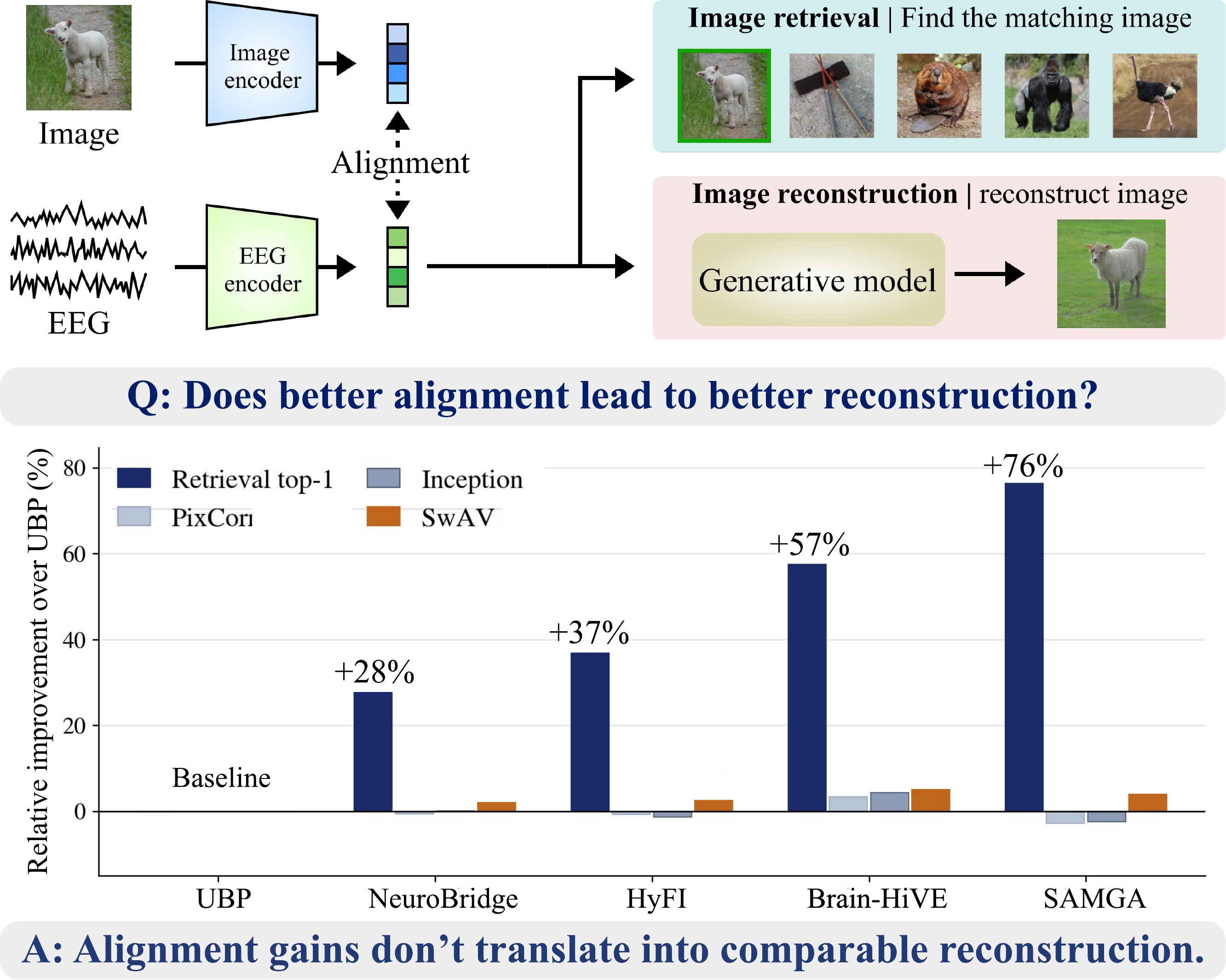}
    \caption{Top: Visually aligned EEG representations are used for both image retrieval and conditioning image reconstruction. Bottom: Relative performance changes across EEG alignment methods using SiT as the generator. Substantial retrieval gains do not translate into comparable reconstruction improvements.}
    \label{fig:motivation}
\end{figure}

\section{Introduction}
\label{sec:intro}

Decoding visual experience from brain activity is a central goal of neuroscience and an important step toward brain-computer
interfaces (BCIs). By translating brain state into human-interpretable visual content, brain decoding can support non-invasive communication while also providing a window into how the brain represents the external world~\cite{naselaris2011encoding, benchetrit2024brain,scotti2024mindeye2}. Among
neural recording modalities, electroencephalography (EEG) is particularly attractive for practical BCI applications because it is non-invasive,
low-cost, portable, and offers high temporal resolution. However, its low
spatial resolution and low signal-to-noise ratio make the recovery of visual information highly challenging. Despite these challenges, recent advances in deep representation learning
and generative modeling have enabled substantial progress in EEG-based
visual decoding, improving both neural representation learning and the
generation of image from EEG signals~\cite{zhang2025cognitioncapturer, daiautoregressive}.

\begin{figure*}[t]
    \centering
    \includegraphics[width=1\linewidth]{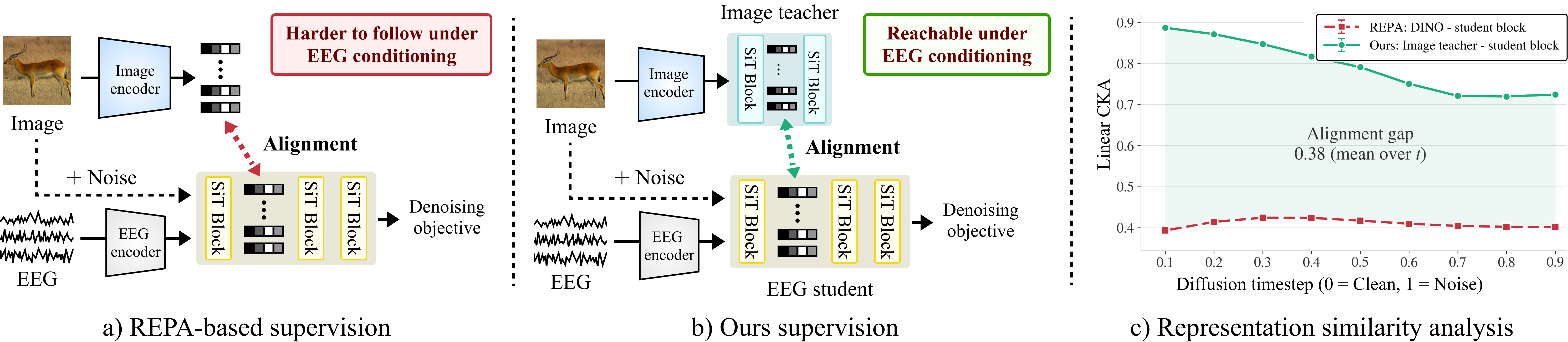}

\caption{Generation-derived supervision for EEG-conditioned reconstruction. a) REPA aligns the generator’s intermediate representations with pretrained image features. b) We instead use intermediate features from an image-conditioned SiT teacher as supervision targets. c) Linear CKA analysis across diffusion timesteps shows stronger alignment with image-teacher features than with image-encoder features.}
    \label{fig:intro2}
\end{figure*}

Recent EEG-based visual decoding research has progressed along two main directions: \emph{EEG-to-image alignment} and \emph{EEG-to-image reconstruction}. EEG-to-image alignment maps neural responses and visual stimuli into a shared representation space and is typically evaluated through image retrieval. Existing approaches have improved alignment by using representations from pretrained vision-language models (VLMs), such as CLIP~\cite{du2023decoding}, blurred-image supervision as in UBP~\cite{wu2025bridging} and NeuroBridge~\cite{zhang2026neurobridge}, hyperbolic feature interpolation in HyFI~\cite{jo2026hyfi}, fused VAE and vision-encoder features in Brain-HiVE~\cite{zhenglearning}, and multi-granularity intermediate features in SAMGA~\cite{jiang2026subject}. Together, these advances highlight perceptually grounded visual features as effective alignment targets and have substantially improved retrieval performance.

In contrast, progress in EEG-to-image reconstruction has been comparatively limited, as reconstruction requires recovering detailed visual content beyond cross-modal alignment. Nevertheless, most existing methods align EEG signals with visual representations and use the resulting embeddings to condition image generators. For instance, several works~\cite{li2024visual, zhang2025cognitioncapturer, zhenglearning} employ SDXL~\cite{podell2024sdxl} with IP-Adapter~\cite{ye2023ip}, whereas \citet{daiautoregressive} fine-tunes a visual autoregressive model conditioned on EEG representations. These methods largely rely on visually aligned EEG representations as conditioning signals for image generation. This raises a fundamental question: \emph{does better EEG-to-image alignment necessarily translate into better reconstruction?}

Our empirical analysis shows that substantial improvements in EEG-to-image retrieval do not translate into comparable gains in image reconstruction (Fig.~\ref{fig:motivation}). We examine this relationship by using representations from recent EEG-to-image alignment methods to condition a Scalable Interpolant Transformer (SiT)~\cite{ma2024sit}, which already outperforms the previous SOTA IP-Adapter-based pipeline. Relative to the baseline condition, retrieval performance improves by up to 76$\%$, but these gains do not translate into comparable improvements in reconstruction metrics. These findings suggest that future progress in EEG-to-image reconstruction should \emph{move beyond conditioning-centric improvements and more directly address the generative process itself.}

To move beyond simply improving input conditioning and explicitly govern image generation, we leverage representation alignment (REPA) frameworks~\cite{yurepresentation}. REPA provides rich visual supervision during denoising by aligning intermediate diffusion transformer (DiT) representations with features from a pretrained vision encoder like DINO~\cite{oquab2024dinov2}. However, under EEG conditioning, these representations remain poorly aligned with DINO features (Fig.~\ref{fig:intro2}), suggesting a mismatch between the supervision targets and the EEG-conditioned generator.

To address this mismatch, we derive supervision from the generative process itself, illustrated Fig.~\ref{fig:intro2} b). Specifically, we first train an image teacher, a SiT that reconstructs the target image conditioned on its DINO embedding, and use its intermediate features as alignment targets for the EEG-conditioned generator. The teacher thus conveys the same visual information as the DINO target, yet expresses it natively within the denoising process rather than as a rigid encoder output. Empirically, the EEG-conditioned generator aligns substantially better with these generation-derived targets, motivating them as supervision for EEG-to-image reconstruction. Our main contributions are summarized as follows:

\begin{itemize}
    \item We expose a critical bottleneck in visual brain decoding: stronger cross-modal alignment does not guarantee better image reconstruction, necessitating a paradigm shift from mere input conditioning to direct supervision of the generative process.

    \item To overcome the inability of EEG-conditioned generators to align with static vision-encoder features, we propose a novel generative distillation framework. An image-conditioned teacher dynamically translates fixed visual features into denoising trajectories, providing highly compatible supervision targets for the EEG-conditioned student.

    \item Our approach establishes a new state-of-the-art in reconstruction fidelity on both THINGS-EEG and THINGS-MEG. Furthermore, representation similarity analyses confirm its mechanistic superiority, and it generalizes robustly across various EEG conditioning methods.

\end{itemize}


%% file: sec/2_rel.tex
\section{Related Work}
\label{sec:rel}

\subsection{EEG-based Visual Brain Decoding}
EEG-based visual decoding aims to identify or reconstruct perceived visual stimuli from EEG signals. A major line of work learns a shared aligned EEG-image representation space\cite{yu2026d,zhou2026eegit}. \citet{songdecoding} applied contrastive learning between EEG signals and pretrained visual features. Later studies adopted blurred-image supervision
~\cite{wu2025bridging,zhang2026neurobridge}, which
\citet{jo2026hyfi} showed induces a perceptual bias in VLM representations. Other approaches combined VAE and semantic features ~\citep{zhenglearning} or aligned EEG with intermediate vision encoder features~\citep{du2026deep,jiang2026subject}. Together, these works highlight the importance of perceptually compatible visual targets and have driven substantial progress in EEG-to-image alignment.

Progress in EEG-to-image reconstruction has been comparatively limited. Most methods condition pretrained generators on visually aligned EEG representations. \citet{li2024visual} and \citet{zhang2025cognitioncapturer} used diffusion priors and IP-Adapter-based generation, while \citet{zhenglearning} pretrained an IP-Adapter conditioning space with visual features suited to EEG alignment. \citet{daiautoregressive} instead explored coarse-to-fine autoregressive generation. These approaches primarily rely on improving the condition supplied to the generator, whereas we directly supervise the generative process through the intermediate representations of an image-conditioned teacher.
\begin{figure*}[t!]
    \centering
    \includegraphics[width=1\linewidth]{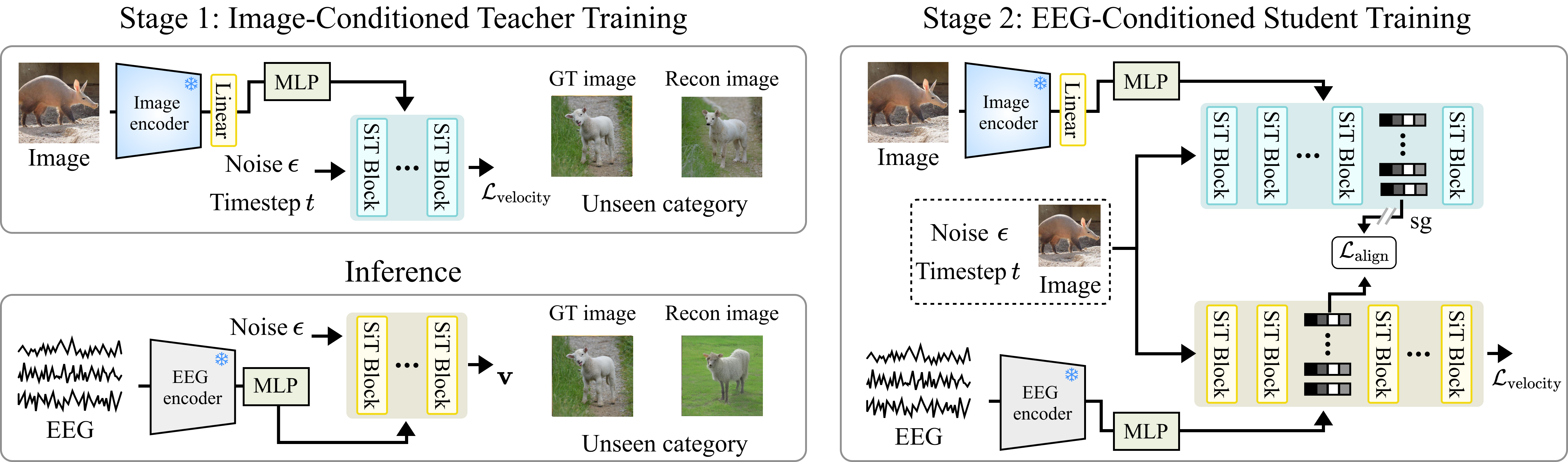}
    \caption{Overall framework of the proposed method, consisting of two training stages. In Stage~1, we train a SiT teacher to reconstruct images conditioned on embeddings from a frozen DINOv2 encoder. In Stage~2, we condition a SiT student on features from a frozen EEG encoder pretrained for EEG-to-image alignment and align the student's intermediate features with those of the frozen teacher. At inference time, only the student and EEG condition are used. $\mathrm{sg}$ denotes the stop-gradient.}
    \label{fig:main}
\end{figure*}

\subsection{Diffusion Models with Representation Alignment}
Diffusion models have become a dominant paradigm for visual generation. Studies analyzing their internal representations reveal discriminative features whose quality closely correlates with generation performance~\cite{li2023your,xiang2023denoising}. Building on this property, REPA~\cite{yurepresentation} introduces additional representation-level supervision during denoising by aligning intermediate diffusion features with those of a pretrained vision encoder such as DINO, improving both training efficiency and generation quality. Subsequent works extend and analyze this paradigm: REG~\cite{wu2026representation} introduces a DINO-aligned representation token directly into the DiT sequence, SRA~\cite{wang2026sra} aligns intermediate features to those of a pretrained VAE, and iREPA~\cite{singh2025matters} shows that the benefits of representation alignment arise primarily from transferring the spatial structure of the target features. Under EEG conditioning, however, we observe that alignment to an external encoder is poorly followed. We therefore use intermediate features from an image-conditioned SiT teacher as supervision targets, which the EEG-conditioned student matches more closely during denoising.

%% file: sec/3_preliminaries.tex
\section{Preliminaries}
\subsection{Flow Matching}
\label{sec:flow_matching}
Our generative model follows the stochastic interpolant formulation~\cite{albergo2025stochastic, ma2024sit}, which provides a unified view of flow- and diffusion-based generative models. Let $\mathbf{x}_* \sim p_{\mathrm{data}}$ be a data sample and $\boldsymbol{\epsilon} \sim \mathcal{N}(\mathbf{0}, \mathbf{I})$ Gaussian noise. We define a time-dependent interpolation between them as
\begin{equation}
\label{eq:interpolation}
    \mathbf{x}_t = \alpha_t \mathbf{x}_* + \sigma_t \boldsymbol{\epsilon}, \qquad t \in [0,T],
\end{equation}
where $\alpha_t$ decreases and $\sigma_t$ increases with $t$, satisfying $\alpha_0=1$, $\sigma_0=0$, $\alpha_T=0$, and $\sigma_T=1$. Consequently, $\mathbf{x}_t$ transitions from clean data at $t=0$ to Gaussian noise at $t=T$.

The marginals induced by Eq.~\eqref{eq:interpolation} can be transported by a probability-flow ordinary differential equation (ODE)
\begin{equation}
\label{eq:velocity_field}
    \frac{\mathrm{d}\mathbf{x}_t}{\mathrm{d}t} = \mathbf{v}(\mathbf{x}_t, t),
\end{equation}
whose time-$t$ marginal coincides with that of the interpolant. Differentiating Eq.~\eqref{eq:interpolation} with respect to $t$ gives the pathwise velocity $\dot{\mathbf{x}}_t = \dot{\alpha}_t \mathbf{x}_* + \dot{\sigma}_t \boldsymbol{\epsilon}$. Since multiple pairs $(\mathbf{x}_*, \boldsymbol{\epsilon})$ can yield the same $\mathbf{x}_t$, the velocity field is given by the conditional expectation
\begin{equation}
\label{eq:velocity_expectation}
    \mathbf{v}(\mathbf{x}, t) = \dot{\alpha}_t\, \mathbb{E}[\mathbf{x}_* \mid \mathbf{x}_t = \mathbf{x}] + \dot{\sigma}_t\, \mathbb{E}[\boldsymbol{\epsilon} \mid \mathbf{x}_t = \mathbf{x}].
\end{equation}

We parameterize this velocity field with a neural network $\mathbf{v}_\theta(\mathbf{x}_t, t)$ and optimize the velocity-matching objective
\begin{equation}
\label{eq:velocity_loss}
    \mathcal{L}_{\mathrm{velocity}}(\theta) = \mathbb{E}_{\mathbf{x}_*, \boldsymbol{\epsilon}, t}\left[\left\| \mathbf{v}_\theta(\mathbf{x}_t, t) - \left(\dot{\alpha}_t \mathbf{x}_* + \dot{\sigma}_t \boldsymbol{\epsilon}\right)\right\|_2^2\right],
\end{equation}
whose minimizer recovers Eq.~\eqref{eq:velocity_expectation}.

Following SiT~\citep{ma2024sit}, we adopt the linear schedule
\begin{equation}
\label{eq:linear_schedule}
    \alpha_t = 1-t, \qquad \sigma_t = t, \qquad T=1,
\end{equation}
for which $\dot{\alpha}_t = -1$ and $\dot{\sigma}_t = 1$, so the velocity target simplifies to $\boldsymbol{\epsilon} - \mathbf{x}_*$.


%% file: sec/4_method.tex
\section{Method}
\subsection{Problem Formulation}
\label{sec:problem}

We address the task of EEG-to-image reconstruction, where the goal is to reconstruct the visual stimulus from the corresponding EEG signal. We formulate this as EEG-conditioned image generation using flow matching.

Let $\mathbf{x}_e \in \mathbb{R}^{C \times L}$ denote a preprocessed EEG signal and $\mathbf{x}_v \in \mathbb{R}^{H \times W \times 3}$ denote its paired visual stimulus, where $C$ and $L$ are the number of EEG channels and time points, and $H$ and $W$ are the image height and width, respectively. The target image is encoded into a latent representation $\mathbf{z}_0=\mathcal{E}(\mathbf{x}_v)$, where $\mathcal{E}$ denotes the encoder of a pretrained visual encoder, such as VAE. The EEG condition is obtained as $\mathbf{c}_e=h_\phi(f_e(\mathbf{x}_e))$, where $f_e$ is an EEG encoder pretrained through EEG-to-image alignment and $h_\phi$ is a learnable projection network introduced for the reconstruction stage. In our setting, we instantiate the data sample as the image latent $\mathbf{z}_v$ and learn an EEG-conditioned velocity field $\mathbf{v}^S_\phi(\mathbf{z}_t, t, \mathbf{c}_e)$ through velocity matching.

\subsection{Overview}
\label{sec:overview}

As illustrated in Fig.~\ref{fig:main}, our framework consists of an image-conditioned teacher $\mathbf{v}^{T}_\theta$ and an EEG-conditioned student $\mathbf{v}^{S}_\phi$ built on the same architecture, where $\theta$ and $\phi$ denote the teacher and student parameters, respectively. The student receives the EEG condition $\mathbf{c}_e$, whereas the teacher is conditioned on image-encoder features.

We first train the teacher once for image-conditioned reconstruction using features from a pretrained visual foundation model, DINOv2~\cite{oquab2024dinov2}, and then reuse the frozen teacher across all subjects. During student training, both models receive the same interpolated latent $\mathbf{z}_t$ and timestep $t$, differing only in their conditioning representations. The student is optimized with the standard conditional velocity-matching objective, while its intermediate representations are additionally aligned with those of the teacher.

At inference, the image-conditioned teacher is discarded. Starting from Gaussian noise $\mathbf{z}_T\sim\mathcal{N}(\mathbf{0},\mathbf{I})$, only the EEG-conditioned student is used to integrate the learned velocity field backward from $t=T$ to $t=0$. The resulting latent $\mathbf{z}_0$ is decoded by the pretrained decoder $\mathcal{D}$ to produce the reconstructed image $\hat{\mathbf{x}}_v=\mathcal{D}(\mathbf{z}_0)$.

\begin{figure*}[t]
    \centering
    \includegraphics[width=1\linewidth]{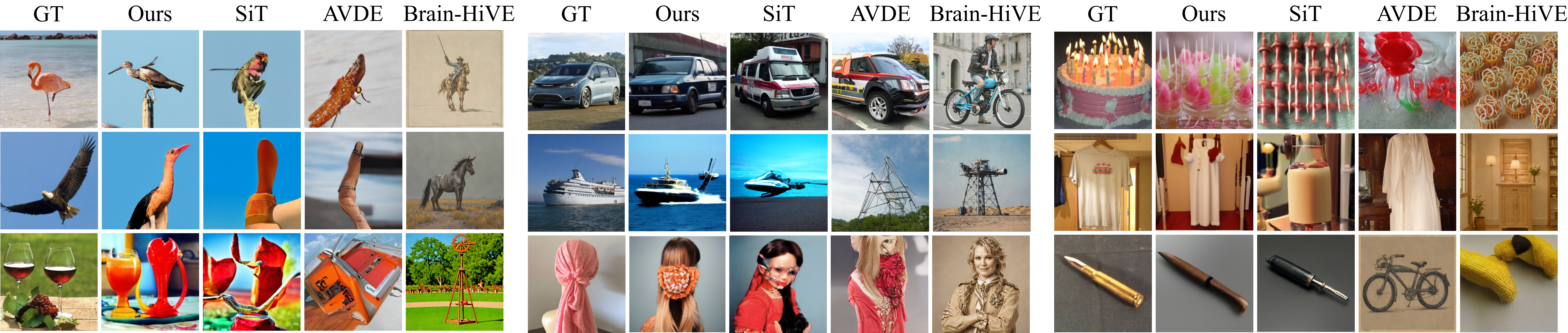}
\caption{Qualitative comparison of EEG-to-image reconstruction on THINGS-EEG (subject 8). Compared with SiT and prior methods, our method more faithfully preserves the category, shape, and color of the stimulus.}
\label{fig:qualitative_r}
\end{figure*}

\subsection{Image-Conditioned Teacher}
\label{sec:image_teacher}
Representation alignment to pretrained vision encoders such as DINO has been shown to improve generation quality of diffusion transformers~\cite{yurepresentation}. However, the EEG-conditioned generator’s intermediate representations remain weakly aligned with DINO features despite this supervision (Fig.~\ref{fig:intro2}).  To provide supervision that the EEG-conditioned generator can follow more closely, we draw alignment targets from a generative model rather than from a vision encoder. To this end, we first train an image-conditioned teacher whose intermediate representations subsequently guide the EEG-conditioned student.

Given a target image $\mathbf{x}_v$, we extract a visual representation 
$\mathbf{r}_v = f_v(\mathbf{x}_v)$ using a frozen DINOv2 encoder $f_v$. A projection head $h_\theta(\cdot)$ maps $\mathbf{r}_v$ to the teacher condition 
$\mathbf{c}_v = h_\theta(\mathbf{r}_v)$, which is injected into every SiT block through adaptive layer normalization (AdaLN), following~\cite{peebles2023scalable}. We zero-initialize the final layer of $h_\theta$, allowing the pretrained generator to adapt gradually.
With the image latent 
$\mathbf{z}_0 = \mathcal{E}(\mathbf{x}_v)$, we sample a timestep $t$ and noise 
$\boldsymbol{\epsilon}$ to form the interpolated latent
\begin{equation}
    \mathbf{z}_t = \alpha_t \mathbf{z}_0 + \sigma_t \boldsymbol{\epsilon},
\end{equation}
and train the teacher $\mathbf{v}^{T}_\theta$ with the conditional velocity matching objective
\begin{equation}
    \mathcal{L}_{\mathrm{teacher}}(\theta)
    = \mathbb{E}_{\mathbf{x}_v, t, \boldsymbol{\epsilon}}
    \left[
    \left\| \mathbf{v}^{T}_\theta(\mathbf{z}_t, t, \mathbf{c}_v) - \mathbf{v}_t^{*} \right\|_2^2
    \right],
\end{equation}
where $\mathbf{v}_t^{*}$ is the target velocity. Because it is conditioned on the image representation, the teacher reconstructs the target image with high fidelity, and its intermediate features encode image-grounded structure (reconstruction results of the image teacher are provided in the Appendix). The teacher is trained once, independently of the subject-specific EEG encoders; it is then frozen and reused across all subjects to supervise the EEG-conditioned student.

\subsection{EEG-Conditioned Student Training}
The student shares the teacher's architecture but is conditioned on EEG rather than on the image. Given an EEG signal $\mathbf{x}_e$, a frozen EEG encoder $f_e$ (pretrained through EEG-to-image alignment) produces $\mathbf{r}_e = f_e(\mathbf{x}_e)$, which the 
projection head maps to the student condition $\mathbf{c}_e = h_{\phi}(\mathbf{r}_e)$. The student $\mathbf{v}^{S}_\phi$ is then trained  with the same velocity-matching objective:
\begin{equation}
    \mathcal{L}_{\mathrm{velocity}}(\phi)
    = \mathbb{E}_{\mathbf{x}_e, t, \boldsymbol{\epsilon}}
    \left[
    \left\| \mathbf{v}^S_\phi(\mathbf{z}_t, t, \mathbf{c}_e) - \mathbf{v}_t^{*} \right\|_2^2
    \right].
\end{equation}

\paragraph{Feature alignment}
To transfer the teacher's generation-oriented representations, we align the student's intermediate features to the teacher's.

Let $\mathbf{H}^{S}_{l}(\mathbf{z}_t, t, \mathbf{c}_e)$ and 
$\mathbf{H}^{T}_{l'}(\mathbf{z}_t, t, \mathbf{c}_v)$ denote the hidden features at the $l$-th block of the student and the $l'$-th block of the teacher, respectively. We align them with
\begin{equation}
    \mathcal{L}_{\mathrm{align}}(\phi)
    = -\mathbb{E}_{\mathbf{x}, \boldsymbol{\epsilon}, t}\!\left[
    \frac{1}{N}\sum_{n=1}^{N}
    \mathrm{sim}\!\left(
    \mathrm{sg}\!\left(\mathbf{H}^{T,[n]}_{l'}\right),\;
    \mathbf{H}^{S,[n]}_{l}
    \right)
    \right],
\end{equation}
where $\mathrm{sim}(\cdot,\cdot)$ is the cosine similarity, $[n]$ indexes tokens, and 
$\mathrm{sg}(\cdot)$ denotes the stop-gradient operation, ensuring the frozen teacher provides a fixed target. Empirically, we align an early student block to a later teacher block because the two locations serve complementary roles. In the image-conditioned teacher, object-level spatial structure becomes more pronounced in later layers, making them suitable as image-grounded supervision targets.  In contrast, applying supervision to an early student layer allows the transferred discriminative information to propagate through the remaining denoising network and influence subsequent generation. This is consistent with REPA, which shows that strengthening representations in earlier layers is particularly effective for generation~\cite{yurepresentation}. We analyze this choice and the layer-wise behavior of the teacher's features in the Appendix.

\paragraph{Training objective}
The student is optimized jointly for reconstruction and alignment:
\begin{equation}
    \mathcal{L}_{\mathrm{student}}(\phi)
    = \mathcal{L}_{\mathrm{velocity}}(\phi) + \lambda \mathcal{L}_{\mathrm{align}}(\phi),
\end{equation}
where $\lambda$ is a balancing weight. As analyzed in Sec.~\ref{sec:experiment}, the EEG-conditioned student aligns more closely with these teacher features than with DINOv2 features, and this guidance translates into improved reconstruction.

%% file: sec/5_experiment.tex
\begin{table*}[t]
\centering
\caption{
Quantitative results of reconstruction quality on THINGS-EEG and THINGS-MEG. We report the mean across all subjects ($\pm$ standard deviation). Bold and underline indicate the best and second-best results within each dataset. $^\dagger$: pretrained SiT-XL/2 fine-tuned without representation alignment. On THINGS-EEG, $^{*}$ marks a significant difference between SiT-XL/2 and Ours (two-sided paired Wilcoxon signed-rank test across subjects, $p<0.05$).
}
\small
\setlength{\tabcolsep}{3.5pt}
\begin{tabular}{llccccccc}
\toprule
\multirow{2}{*}{Dataset}
& \multirow{2}{*}{Method}
& \multicolumn{2}{c}{Low-level}
& \multicolumn{5}{c}{High-level} \\
\cmidrule(lr){3-4}
\cmidrule(lr){5-9}
&
& PixCorr $\uparrow$
& SSIM $\uparrow$
& AlexNet(2) $\uparrow$
& AlexNet(5) $\uparrow$
& Inception $\uparrow$
& CLIP $\uparrow$
& SwAV $\downarrow$ \\
\midrule

\multirow{4}{*}{MEG}
& AVDE
& $0.124 \pm .017$
& $\mathbf{0.339} \pm .009$
& $0.728 \pm .009$
& $0.801 \pm .007$
& $0.685 \pm .012$
& $0.720 \pm .016$
& $0.651 \pm .006$ \\

& Brain-HiVE
& $0.122 \pm .016$
& $0.236 \pm .021$
& $0.672 \pm .046$
& $0.772 \pm .051$
& $0.656 \pm .040$
& $0.704 \pm .025$
& $0.638 \pm .021$ \\
\cmidrule(l){2-9}

& \textit{SiT-XL/2}$^\dagger$
& $\underline{0.155} \pm .050$
& $0.296 \pm .028$
& $\underline{0.791} \pm .070$
& $\underline{0.864} \pm .065$
& $\underline{0.755} \pm .072$
& $\underline{0.744} \pm .059$
& $\underline{0.565} \pm .033$ \\

\rowcolor{gray!12}
& \textbf{Ours}
& $\mathbf{0.162} \pm .046$
& $\underline{0.328} \pm .030$
& $\mathbf{0.800} \pm .075$
& $\mathbf{0.874} \pm .053$
& $\mathbf{0.762} \pm .054$
& $\mathbf{0.745} \pm .042$
& $\mathbf{0.557} \pm .030$ \\

\midrule
\midrule

\multirow{5}{*}{EEG}
& Cog-Cap & $0.150 \pm .013$ & $0.347 \pm .022$ & $0.754 \pm .026$ & $0.623 \pm .032$ & $0.669 \pm .043$ & $0.715 \pm .031$ & $0.590 \pm .025$ \\
& AVDE & $0.147 \pm .023$ & $\underline{0.366} \pm .018$ & $0.766 \pm .025$ & $0.835 \pm .026$ & $0.724 \pm .025$ & $0.747 \pm .030$ & $0.586 \pm .021$ \\
& Brain-HiVE & $0.195 \pm .013$ & $0.336 \pm .014$ & $0.843 \pm .016$ & $0.905 \pm .015$ & $0.756 \pm .025$ & $\mathbf{0.808} \pm .025$ & $0.554 \pm .015$ \\
\cmidrule(l){2-9}
& \textit{SiT-XL/2}$^\dagger$ & $\underline{0.261} \pm .024$ & $0.346 \pm .021^{*}$ & $\underline{0.916}\pm .014^*$ & $\underline{0.941} \pm .011$ & $\underline{0.799} \pm .031^{*}$ & $0.791 \pm .032^{*}$ & $\underline{0.508} \pm .019^{*}$ \\
\rowcolor{gray!12}
& \textbf{Ours} & $\mathbf{0.263} \pm .026$ & $\mathbf{0.370} \pm .020$ & $\mathbf{0.921} \pm .016$ & $\mathbf{0.942} \pm .012$ & $\mathbf{0.813} \pm .019$ & $\underline{0.799} \pm .029$ & $\mathbf{0.500} \pm .020$ \\

\bottomrule
\end{tabular}
\label{tab:main}
\end{table*}

\section{Experiments}
\label{sec:experiment}

\subsection{Datasets}
\paragraph{THINGS-EEG}
We conduct our experiments on THINGS-EEG~\cite{gifford2022large}, a large-scale visual EEG dataset acquired using a rapid serial visual presentation paradigm (RSVP). The dataset contains paired image and EEG recordings from 10 participants. Its training split covers 1,654 object concepts with 10 images per concept, where each image was presented four times to each participant. The test split contains new 200 concepts represented by one image each, with 80 repetitions per participant. Following established protocols~\cite{wu2025bridging,zhenglearning}, we average repeated trials, resample the EEG recordings to 250~Hz, and retain 17 electrodes located over the occipital and parietal regions.

\paragraph{THINGS-MEG}
We additionally evaluate our approach on THINGS-MEG~\cite{hebart2023things}, which provides visual MEG recordings from four participants using 271 sensors. The training split comprises 1,854 object concepts, each represented by 12 distinct images presented once. The test split includes 200 concepts with one image per concept, repeated 12 times. We apply the preprocessing protocol used in prior neural decoding studies ~\cite{wu2025bridging,zhenglearning}; further implementation details are provided in the appendix.

\subsection{Implementation details}
We initialize both the teacher and student from an ImageNet-pretrained SiT-XL/2 checkpoint~\cite{ma2024sit} and fully fine-tune each backbone during its respective training stage. We train the image-conditioned teacher using only images from the THINGS training split and then freeze it for student training. By default, the student uses EEG embeddings from Brain-HiVE~\cite{zhenglearning}. A conditioning MLP projects each embedding to 1152 dimensions and supplies it to every SiT block through adaptive layer normalization (AdaLN), following DiT~\cite{peebles2023scalable}. We zero-initialize the final layer of the MLP so that the EEG conditioning signal is introduced gradually during training. For classifier-free guidance, we retain the null embedding from the pretrained checkpoint. We align student block~8 with teacher block~25 using a token-wise cosine loss weighted by $\lambda=0.3$. The blocks and loss weight are selected on a validation set formed by holding out 10\% of the training split. The Appendix details how we select the blocks and loss weight. We train each model for 200 epochs using AdamW with a learning rate of $3\times10^{-5}$ and a batch size of 64 on a single NVIDIA B200 GPU. At inference, we use an Euler ODE solver with 50 NFE and a classifier-free guidance scale of 5.

\subsection{Result}
To evaluate decoding performance, we perform EEG-to-image reconstruction on THINGS-EEG and THINGS-MEG. Following prior work~\cite{zhenglearning,daiautoregressive}, we compare reconstructions with ground-truth images using pixel-wise correlation (PixCorr) and structural similarity (SSIM) for low-level fidelity. For high-level similarity, we report two-way identification scores using AlexNet (layers 2 and 5), Inception, and CLIP, along with SwAV distance (lower is better).
We compare our method with recent EEG-to-image reconstruction
approaches. Cog-Cap~\cite{zhang2025cognitioncapturer} aligns EEG
with image, text, and depth representations and reconstructs images using a diffusion prior with pretrained SDXL and IP-Adapters. AVDE~\cite{daiautoregressive} adopts a visual autoregressive model conditioned on EEG representations. Brain-HiVE~\cite{zhenglearning} pretrains an IP-Adapter using fused visual representations and subsequently aligns EEG embeddings with the learned conditioning space for image generation. Results of these methods in Table~\ref{tab:main} are taken from their original papers.

As shown in Table~\ref{tab:main}, our method achieves the best mean performance on six of seven metrics on both THINGS-EEG and THINGS-MEG. We compare against results reported in prior work and a SiT-XL/2 baseline trained on the same Brain-HiVE embeddings without representation alignment. On THINGS-EEG, our method improves over SiT-XL/2 on all seven metrics. PixCorr increases from 0.261 to 0.263, SSIM from 0.346 to 0.370, and Inception from 0.799 to 0.813. Among these improvements, four metrics show statistically significant gains across subjects (two-sided paired Wilcoxon signed-rank test, $p<0.05$). Our method also achieves the highest AlexNet(2) score (0.921) and the lowest SwAV distance (0.500), while Brain-HiVE achieves the highest CLIP score (0.808).

On THINGS-MEG, our method achieves the highest PixCorr (0.162) and the best scores on all five high-level metrics. SSIM improves from 0.296 with SiT-XL/2 to 0.328 with our method, below AVDE's 0.339. Because THINGS-MEG includes only four subjects, we report descriptive results without significance testing. Fig.~\ref{fig:qualitative_r} shows reconstructions for THINGS-EEG subject~8, including examples in which our method better preserves the category, shape, and color of the target image. Additional per-subject results are provided in the Appendix.

\begin{table}[t]
\centering
\caption{Comparison of teacher transfer strategies on THINGS-EEG (mean over 10 subjects). All methods use the same frozen image-conditioned teacher and differ only in where its knowledge is transferred. Bold denotes the best result.}
\small
\setlength{\tabcolsep}{3.4pt}
\begin{tabular}{llcccc}
\toprule
\multirow{2}{*}{Method} & \multirow{2}{*}{Supervises} & \multicolumn{2}{c}{Low-level} & \multicolumn{2}{c}{High-level} \\
\cmidrule(lr){3-4} \cmidrule(lr){5-6}
 & & PixCorr$\uparrow$ & SSIM$\uparrow$ & CLIP$\uparrow$ & SwAV$\downarrow$ \\
\midrule
Condition align. & Input   & 0.228 & \underline{0.365} & 0.782 & 0.513 \\
Velocity distill. & Output  & \textbf{0.264} & 0.353 & \underline{0.790} & \underline{0.506} \\
\rowcolor{gray!12}
\textbf{Ours}     & Features & \underline{0.263} & \textbf{0.370} & \textbf{0.799} & \textbf{0.500} \\
\bottomrule
\end{tabular}
\label{tab:transfer}
\end{table}

\subsection{Ablation Study}

\paragraph{Comparison of Teacher Transfer Strategies}
We examine where the teacher's knowledge should be transferred by comparing our feature alignment with two alternatives that use the same frozen teacher at different stages of the model (Table~\ref{tab:transfer}). Specifically, we distill the teacher at three points of the generator: its input condition, its intermediate representations (ours), and its final velocity output. Condition alignment applies a contrastive objective between the EEG and image conditions before they enter the generator, while velocity distillation matches the student's velocity prediction to the teacher's. Condition alignment performs worst, particularly on low-level metrics, indicating that making the EEG condition resemble the image condition does not by itself improve reconstruction. Velocity distillation is more competitive and matches ours in PixCorr, but falls behind in SSIM and the high-level metrics. These results show that the teacher is most effective when it supervises the intermediate representations of the generator, rather than its input condition or its output. Detailed results are provided in Appendix.

\begin{figure}[t]
  \centering
  \includegraphics[width=\linewidth]{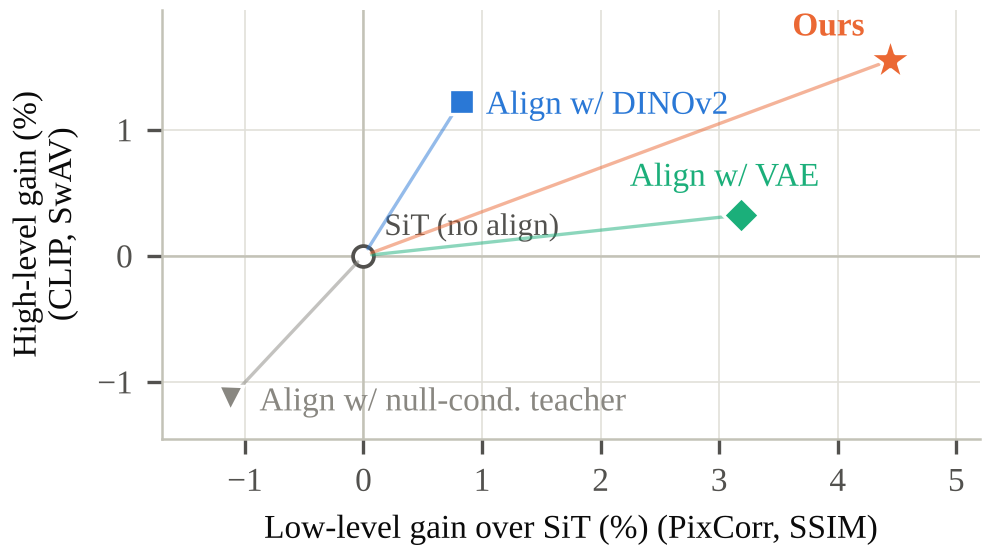}
    \caption{Ablation on the alignment target on THINGS-EEG. Each point shows the mean relative improvement over the unaligned SiT in low-level (PixCorr, SSIM) and high-level (CLIP, SwAV; sign flipped for SwAV) metrics, averaged over 10 subjects.}
  \label{fig:ablation_target}
\end{figure}

\paragraph{Comparison of Alignment Target}
We examine how the choice of alignment target affects reconstruction. Under the same EEG-conditioned SiT backbone and EEG condition, we compare five settings: no additional alignment (SiT), alignment to DINOv2 features following~\cite{yurepresentation} (Align w/ DINOv2), alignment to VAE features following~\cite{wang2026sra} (Align w/ VAE), alignment to an unconditional teacher that receives no image condition (Align w/ uncond.\ teacher), and alignment to the proposed image-conditioned teacher (Ours). This comparison isolates the effect of the supervision target.

As shown in Fig.~\ref{fig:ablation_target}, we select two representative metrics for each level, PixCorr and SSIM for low-level and CLIP and SwAV for high-level reconstruction, and measure the average relative gain over SiT within each level. The two external targets each improve only one side of reconstruction: aligning to DINOv2 features yields a clear high-level gain but little low-level gain, whereas aligning to VAE features shows the opposite pattern, improving low-level metrics with little high-level gain. Aligning to the same teacher under the null condition degrades both levels below SiT, indicating that aligning to generator features alone is not beneficial. In contrast, aligning to the image-conditioned teacher improves both, achieving the largest low-level gain while matching the high-level gain of DINOv2 alignment. Notably, the teacher is conditioned on the same DINOv2 features, indicating that the same visual information becomes an effective target for both low- and high-level reconstruction once it is expressed within the generator rather than provided as a fixed encoder output. Detailed results on all metrics are provided in the Appendix.

\begin{table}[t]
\centering
\caption{Reconstruction on THINGS-EEG under three EEG conditioning representations obtained from EEG-to-image alignment methods. All models are trained for 100 epochs.}
\label{tab:eeg_conditions}
\setlength{\tabcolsep}{3pt}
\small
\resizebox{\columnwidth}{!}{%
\begin{tabular}{@{}llrrrr@{}}
\toprule
Condition & Generator & PixCorr $\uparrow$ & SSIM $\uparrow$ & CLIP $\uparrow$ & SwAV $\downarrow$ \\
\midrule
\multirow{4}{*}{UBP}
 & IP-Adapter & $0.184$ & $\mathbf{0.379}$ & $0.742$ & $0.601$ \\
 & AVDE & $0.160$ & $0.369$ & $0.742$ & $0.545$ \\
 & SiT & $0.236$ & $0.330$ & $0.757$ & $0.547$ \\
\rowcolor{gray!12}
\cellcolor{white} & Ours & $\mathbf{0.245}$ & $0.363$ & $\mathbf{0.802}$ & $\mathbf{0.508}$ \\
\midrule
\multirow{4}{*}{Brain-HiVE}
 & IP-Adapter & $0.188$ & $0.355$ & $0.724$ & $0.603$ \\
 & AVDE & $0.164$ & $0.365$ & $0.736$ & $0.546$ \\
 & SiT & $0.261$ & $0.360$ & $0.787$ & $0.516$ \\
\rowcolor{gray!12}
\cellcolor{white} & Ours & $\mathbf{0.263}$ & $\mathbf{0.377}$ & $\mathbf{0.792}$ & $\mathbf{0.513}$ \\
\midrule
\multirow{4}{*}{SAMGA}
 & IP-Adapter & $0.193$ & $0.351$ & $0.744$ & $0.582$ \\
 & AVDE & $0.161$ & $0.367$ & $0.730$ & $0.537$ \\
 & SiT & $0.217$ & $0.357$ & $0.789$ & $0.522$ \\
\rowcolor{gray!12}
\cellcolor{white} & Ours & $\mathbf{0.257}$ & $\mathbf{0.384}$ & $\mathbf{0.798}$ & $\mathbf{0.511}$ \\
\bottomrule
\end{tabular}
}
\end{table}


\subsection{Generalization Across EEG Conditions}
We evaluate whether the benefit of generation-derived supervision extends across heterogeneous EEG conditioning representations on all 10 THINGS-EEG subjects, using embeddings from UBP~\cite{wu2025bridging} (blurred-image supervision), Brain-HiVE~\cite{zhenglearning} (fused VAE and vision-encoder features), and SAMGA~\cite{jiang2026subject} (intermediate InternViT features). For each condition, we compare against an IP-Adapter-based generator~\cite{zhenglearning} and the autoregressive AVDE~\cite{daiautoregressive} (see Table~\ref{tab:generative_comparison}), as well as a SiT baseline with the same backbone but without representation alignment, to isolate the effect of generation-derived supervision. All models are trained for 100 epochs.

As shown in Table~\ref{tab:eeg_conditions}, GEARS consistently improves over SiT on all four metrics under every condition. Among all generators, it achieves the best PixCorr, CLIP, and SwAV scores under every condition, and the best SSIM with Brain-HiVE and SAMGA. Notably, the largest gains occur where SiT performs weakest, reducing the cross-condition range across all four metrics. Thus, once the generator is supervised, reconstruction depends less on the choice of EEG embedding, supporting generator-level supervision as complementary to alignment.



\subsection{Analysis}


\begin{figure}
    \centering
    \includegraphics[width=0.9\linewidth]{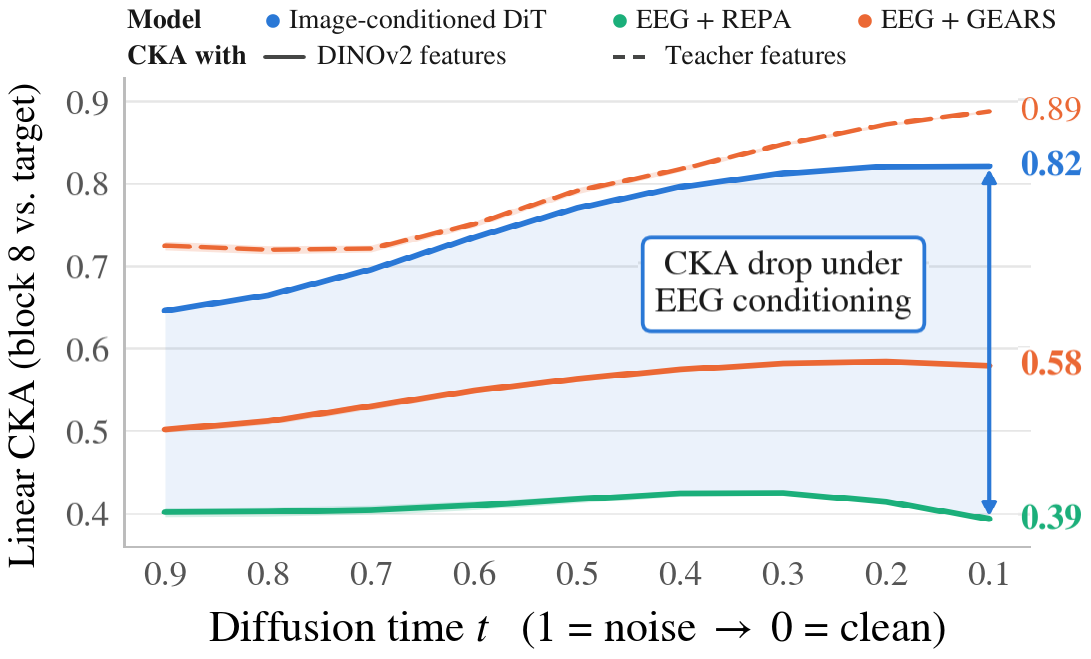}
\caption{Linear CKA between block~8 features and DINOv2 (solid) or teacher features (dashed) across diffusion timesteps. GEARS aligns more closely with DINOv2.}
\label{fig:cka_block8}
\end{figure}

\paragraph{Representation Alignment During Denoising}
Building on Fig.~\ref{fig:intro2}, we examine how the choice of supervision target shapes the EEG-conditioned student's representations during denoising. At each diffusion timestep, we compute linear centered kernel alignment (CKA)~\cite{kornblith2019similarity} between the student's block~8 features and each reference representation (DINOv2 or image-conditioned teacher features), then average the scores across all 10 THINGS-EEG subjects.

As shown in Fig.~\ref{fig:cka_block8}, the image-conditioned DiT aligns closely with DINOv2 (blue), whereas the EEG-conditioned DiT shows weaker alignment even under direct DINOv2 supervision (green). The teacher-supervised EEG model closely follows the image-conditioned teacher throughout denoising (dashed orange). It also aligns more closely with DINOv2 than the directly supervised model at every evaluated timestep (solid orange vs.\ green). This suggests that the image-conditioned teacher conveys visual features in a form the EEG-conditioned model can more readily learn.

\paragraph{Layer-wise Representation Analysis}
To examine how teacher supervision shapes the denoising network, we compute the CKA between DINO features and the hidden representations of each DiT block. Unlike Fig.~\ref{fig:cka_block8}, features are extracted along the EEG-conditioned sampling trajectory at inference, without access to the ground-truth latent, and averaged over all subjects on the THINGS-EEG test set. As shown in Fig.~\ref{fig:layer}, our model achieves consistently higher DINO similarity in the early-to-intermediate blocks than both vanilla SiT and direct DINO alignment, indicating that the teacher induces more visually structured representations even without DINO as the supervision target. In contrast, direct DINO alignment yields lower similarity across much of the network and only limited gains in reconstruction, suggesting that teacher supervision shapes visual representations beyond the directly supervised block.



%% file: sec/6_conclusion.tex
\begin{figure}[t]
    \centering
    \includegraphics[width=1\linewidth]{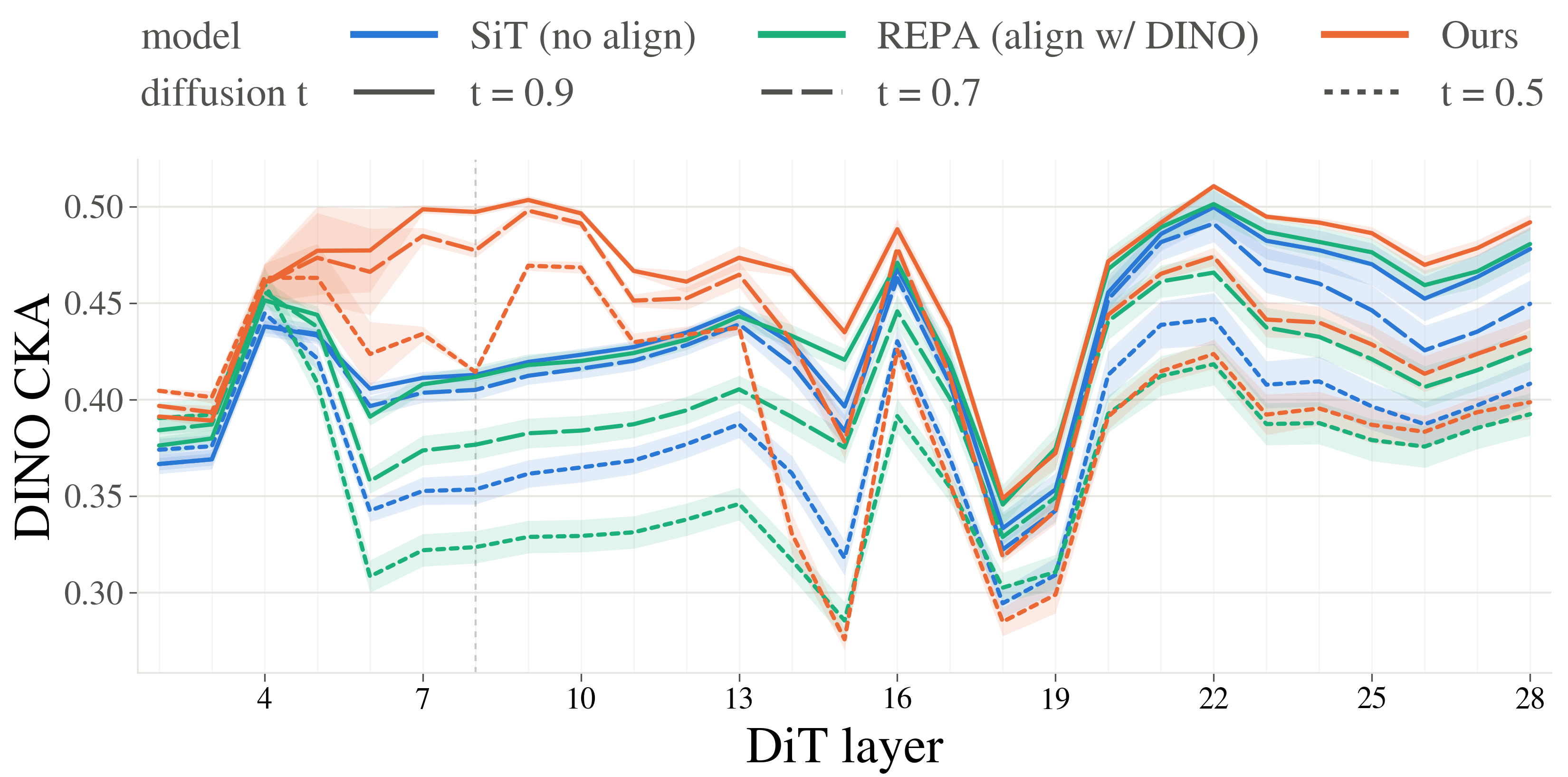}
    \caption{
    Analysis of layer-wise DINO CKA across the DiT at three diffusion timesteps. Across all models, our model (orange) attains the highest DINO CKA at early-to-intermediate layers, suggesting that the discriminative nature of these intermediate features contributes to the improved reconstruction.
    }
\label{fig:layer}
\end{figure}

\section{Conclusion}
\label{sec:concul}

We showed that substantial EEG-to-image retrieval gains do not necessarily translate into comparable reconstruction improvements, and that EEG-conditioned generators struggle to align with DINOv2 even under direct supervision. To address this, we introduced GenErAtion-deRived Supervision (GEARS), which supervises denoising with intermediate features from an image-conditioned diffusion teacher. The student closely follows these targets and achieves stronger DINOv2 alignment than direct supervision. GEARS consistently improves upon a strong SiT baseline across three EEG representations, with larger gains for weaker baselines, suggesting reduced sensitivity to EEG conditioning. It achieves state-of-the-art reconstruction on THINGS-EEG and THINGS-MEG, establishing generator-level supervision as a complementary direction to EEG-to-image alignment.

\paragraph{Limitations}
EEG-to-image retrieval gains do not consistently translate into better reconstruction. Which visual cues support image discrimination in retrieval and preserve appearance in reconstruction remains an open question. In addition, the generator is still fully fine-tuned for each individual. We leave lightweight adaptation to new participants with subject-invariant EEG representations to future work.

\section{LLM Usage Statement}
OpenAI ChatGPT and Anthropic Claude were used under author direction to assist with manuscript editing, the implementation and debugging of experimental and analysis code, and the presentation of results. The authors directed all research questions and experimental procedures, executed the experiments, and reviewed and verified all AI-assisted text, code, citations, results, and interpretations. All scientific claims and decisions are the responsibility of the authors.

%% file: sec/X_suppl.tex
\clearpage
\onecolumn

\setcounter{page}{1}
\appendix
\setcounter{figure}{0}
\setcounter{table}{0}
\setcounter{equation}{0}
\renewcommand{\thefigure}{S\arabic{figure}}
\renewcommand{\thetable}{S\arabic{table}}
\renewcommand{\theequation}{S\arabic{equation}}
\renewcommand{\thesection}{S\arabic{section}}

\onecolumn
\etocdepthtag.toc{supp}
\etocsettagdepth{mainpart}{none}
\etocsettagdepth{supp}{subsection}

\etocsettocstyle{
  \section*{\Large Beyond Alignment: Supervising the Generator for EEG-to-Image Reconstruction}
  \subsection*{Supplementary Material}
  \subsubsection*{Contents}
  \vspace{6pt}
}{}

\etocsetstyle{section}
  {}
  {\par\vspace{8pt}\leftskip0pt\noindent}
  {{\large\bfseries
    \etocthelinkednumber.\ \etocthelinkedname
    \hfill\etocthelinkedpage}\par}
  {}

\etocsetstyle{subsection}
  {}
  {\par\vspace{3pt}\leftskip1.6em\noindent}
  {{\normalsize
    \etocthelinkednumber.\ \etocthelinkedname
    \leaders\hbox to .6em{\hss.\hss}\hfill
    \etocthelinkedpage}\par}
  {}

\tableofcontents

\twocolumn

\begin{table*}[t]
\centering
\caption{Sequential hyperparameter selection on a validation split held out from 10\% of the training data. The best setting on each axis is fixed for the next: $\lambda \rightarrow L \rightarrow t$. Bold denotes the best result within each axis.}
\setlength{\tabcolsep}{4pt}
\begin{tabular}{llccccccccc}
\toprule
\multirow{2}{*}{Axis} & \multirow{2}{*}{Setting}
& \multicolumn{2}{c}{Low-level} & \multicolumn{5}{c}{High-level}
& \multirow{2}{*}{Avg.\ rank} & \multirow{2}{*}{Rank} \\
\cmidrule(lr){3-4} \cmidrule(lr){5-9}
& & PixCorr$\uparrow$ & SSIM$\uparrow$ & AlexNet(2)$\uparrow$ & AlexNet(5)$\uparrow$ & Inception$\uparrow$ & CLIP$\uparrow$ & SwAV$\downarrow$ & & \\
\midrule
\multirow{3}{*}{\shortstack[l]{$\lambda$}}
& \textbf{0.3} & 0.194 & 0.411 & \textbf{0.873} & \textbf{0.901} & \textbf{0.788} & \textbf{0.739} & \textbf{0.547} & \textbf{1.43} & 1 \\
& 0.5 & \textbf{0.208} & \textbf{0.412} & 0.872 & 0.899 & 0.781 & 0.734 & 0.553 & 2.00 & 2 \\
& 0.7 & 0.200 & 0.407 & 0.871 & 0.900 & 0.777 & 0.732 & 0.552 & 2.57 & 3 \\
\midrule
\multirow{3}{*}{\shortstack[l]{$L$}}
& 23 & \textbf{0.205} & \textbf{0.413} & 0.864 & 0.883 & 0.777 & 0.724 & 0.556 & 1.86 & 2 \\
& \textbf{25} & 0.194 & 0.411 & \textbf{0.873} & \textbf{0.901} & \textbf{0.788} & \textbf{0.739} & \textbf{0.547} & \textbf{1.43} & 1 \\
& 27 & 0.199 & 0.404 & 0.860 & 0.891 & 0.765 & 0.722 & 0.558 & 2.71 & 3 \\
\midrule
\multirow{4}{*}{\shortstack[l]{$t$}}
& \textbf{[0, 1]} & 0.194 & 0.411 & \textbf{0.873} & \textbf{0.901} & \textbf{0.788} & \textbf{0.739} & \textbf{0.547} & \textbf{1.57} & 1 \\
& [0.1, 0.4] & \textbf{0.207} & \textbf{0.418} & 0.864 & 0.883 & 0.762 & 0.728 & 0.557 & 2.57 & 2 \\
& [0.4, 0.7] & 0.203 & 0.408 & 0.853 & 0.893 & 0.769 & 0.727 & 0.551 & 2.86 & 3 \\
& [0.7, 1.0] & 0.196 & 0.393 & 0.860 & 0.898 & 0.775 & 0.720 & 0.552 & 3.00 & 4 \\
\bottomrule
\end{tabular}

\label{tab:hparam}
\end{table*}

\section{Datasets}
We evaluate on two publicly available brain-image datasets, THINGS-EEG and THINGS-MEG, 
each pairing neural recordings with the visual stimuli that elicited them. In both, 
stimuli were shown under a Rapid Serial Visual Presentation (RSVP) protocol, in which 
images appear in quick succession while neural activity is recorded with high temporal 
resolution.

\subsection{THINGS-EEG}
THINGS-EEG~\cite{gifford2022large} provides EEG recordings from 10 participants, captured 
with a 64-channel montage sampled at 1000~Hz and band-limited to $[0.1, 100]$~Hz. Its 
training portion spans 1,654 object concepts with 10 images each, and every image is 
shown four times per participant, giving $1{,}654 \times 10 \times 4$ trials. The test 
portion covers a disjoint set of 200 concepts, one image per concept, each repeated 80 
times ($200 \times 80$ trials).

For consistency with earlier studies~\cite{songdecoding, wu2025bridging}, we adopt their 
preprocessing. We resample the signals to 250~Hz and keep the 17 occipito-parietal 
electrodes that overlie visual cortex. Trials sharing the same stimulus are averaged to 
improve the signal-to-noise ratio, leaving 16,540 training and 200 test examples per 
participant.

\subsection{THINGS-MEG}
THINGS-MEG~\cite{hebart2023things} supplies MEG recordings from four participants, 
measured with a 271-sensor array. On each trial an image was displayed for 500~ms and 
followed by a blank inter-stimulus interval of $1000 \pm 200$~ms. Training covers 1,854 
concepts, each shown as 12 distinct images presented a single time, while the test set 
holds 200 held-out concepts, each with one image repeated 12 times.

Following the same prior protocol~\cite{songdecoding, wu2025bridging}, we band-pass filter the MEG between 0.1 and 100~Hz, downsample to 200~Hz. The 200 test concepts are held out from training, as in previous work.

\section{Hyperparameter Selection}
\label{app:hparam}

We select the alignment weight $\lambda$, the teacher layer $L$, and the timestep range of the alignment loss on a validation set formed by randomly holding out one of the ten images of each training concept (10\% of the THINGS-EEG training split), without access to the test set. To reduce computational cost, each configuration is trained for 100 epochs on five randomly selected subjects, and the results are averaged over these subjects. The student layer is fixed to block 8 throughout. We rank the configurations on each of the seven metrics and select the one with the best average rank.

The hyperparameters are selected sequentially (Table~\ref{tab:hparam}). We first select $\lambda=0.3$ with the teacher layer fixed to 25 and the alignment loss applied over the full timestep range. With $\lambda=0.3$, we then select teacher layer 25 over the neighboring layers 23 and 27. Finally, we compare restricting the alignment loss to a sub-range of timesteps ($t=1$: noise, $t=0$: clean) and find that applying it over the full range performs best. Increasing $\lambda$ from 0.3 to 0.5 or using an earlier teacher
layer slightly improves the low-level metrics but degrades
all five high-level metrics.

\begin{table*}[t]
\centering
\caption{
Retrieval accuracy and reconstruction performance across EEG
representations. Bold and underline indicate the best and second-best
results, respectively. Retrieval accuracy is evaluated using the same
EEG encoder checkpoints as reconstruction in our reimplementation.
No alignment denotes an EEG encoder trained from scratch using only
the reconstruction objective; its results are averaged over five seeds.
}
\label{tab:retrieval_recon}
\setlength{\tabcolsep}{2.5pt}
\begin{tabular}{lcccccccc}
\toprule
\multirow{2}{*}{Model}
& \multirow{2}{*}{Retrieval ACC (\%) $\uparrow$}
& \multicolumn{2}{c}{Low-level}
& \multicolumn{5}{c}{High-level} \\
\cmidrule(lr){3-4}
\cmidrule(lr){5-9}
&
& PixCorr $\uparrow$
& SSIM $\uparrow$
& AlexNet(2) $\uparrow$
& AlexNet(5) $\uparrow$
& Inception $\uparrow$
& CLIP $\uparrow$
& SwAV $\downarrow$ \\
\midrule
No alignment
& --
& \underline{0.247}
& 0.354
& 0.880
& 0.908
& 0.762
& 0.759
& 0.545 \\

UBP
& 49.8
& 0.238
& 0.330
& 0.893
& 0.926
& 0.787
& 0.758
& 0.547 \\

NeuroBridge
& 63.9
& 0.231
& 0.332
& 0.892
& 0.928
& \underline{0.793}
& 0.776
& 0.533 \\

HyFI
& 68.3
& 0.230
& 0.308
& \underline{0.895}
& \underline{0.933}
& 0.790
& 0.772
& 0.531 \\

Brain-HiVE
& \underline{78.4}
& \textbf{0.261}
& \textbf{0.360}
& \textbf{0.910}
& \textbf{0.935}
& \textbf{0.803}
& \underline{0.787}
& \textbf{0.516} \\

SAMGA
& \textbf{88.1}
& 0.217
& \underline{0.357}
& 0.893
& 0.924
& 0.790
& \textbf{0.789}
& \underline{0.522} \\
\bottomrule
\end{tabular}
\end{table*}

\section{Implementation Details}
\subsection{Architecture and Training}
\label{sec:arch}

\paragraph{Backbone.}
We build on SiT-XL/2~\cite{ma2024sit}, operating in the latent space of a frozen
Stable Diffusion VAE (\texttt{sd-vae-ft-mse}); a $256\!\times\!256$ image is encoded to a
$32\!\times\!32\!\times\!4$ latent and tokenized with patch size $2$, giving $256$ tokens of
width $1152$ across $28$ blocks. All $677$M backbone parameters are initialized from the
official SiT-XL/2 $256\!\times\!256$ checkpoint and fully fine-tuned; training from scratch or from a partially loaded
checkpoint degrades every metric we report. We train a separate image-conditioned teacher for each dataset, using only the images of its training split, and then freeze it for student training.

\paragraph{Conditioning.}
EEG is encoded by a frozen, subject-specific Brain-HiVE\cite{zhenglearning} encoder into a $1024$-d vector,
which replaces the class embedding consumed by the backbone's adaLN modulation. The
adapter is a lightweight MLP, $\mathrm{LN}(1024)\!\rightarrow\!\mathrm{Linear}(1024,1152)
\!\rightarrow\!\mathrm{SiLU}\!\rightarrow\!\mathrm{Linear}(1152,1152)$. Its final linear layer is zero-initialized, allowing the EEG conditioning signal to be introduced gradually as a learned residual while preserving the pretrained SiT backbone at initialization. For classifier-free guidance we keep a learnable
null embedding initialized from the null entry of the pretrained class-embedding table.
The Brain-HIVE encoder is frozen throughout; only the backbone
and the adapter are updated.

\paragraph{Optimization.}
We use AdamW with $\beta=(0.9,0.95)$ and zero weight decay, a constant
learning rate of $3\!\times\!10^{-5}$ without warmup or decay, a batch
size of $64$, and bfloat16 autocast. Each subject provides $16{,}540$
EEG-image training pairs, so training for $200$ epochs corresponds to
approximately $51.6$K optimization steps. The image-conditioned teacher
is trained using the same optimization configuration.

\section{Relationship Between Retrieval and Reconstruction}

\subsection{Evaluation Setup}
To examine whether stronger EEG-to-image retrieval translates into
better reconstruction, we compare EEG representations from
UBP~\cite{wu2025bridging}, NeuroBridge~\cite{zhang2026neurobridge},
HyFI~\cite{jo2026hyfi}, Brain-HiVE~\cite{zhenglearning}, and
SAMGA~\cite{jiang2026subject} using the same reconstruction architecture and a fixed training budget of 100 epochs.
For Brain-HiVE, we use the configuration that fuses VAE, CLIP-B, and SynCLR~\cite{tian2024learning} features, which attains a retrieval accuracy of 78.4\%. We select this configuration because its accuracy lies between those of HyFI (68.3\%) and SAMGA (88.1\%), so that the compared representations cover the retrieval range more evenly and the relationship between retrieval and reconstruction can be examined more clearly.
We additionally include a \emph{no-alignment} control, in which the EEG
encoder is trained from scratch using only the reconstruction objective, without EEG-to-image alignment pretraining.
For this control, we report results averaged over five seeds.

\subsection{Analysis}
As shown in Table~\ref{tab:retrieval_recon}, retrieval accuracy
ranges from $49.8\%$ to $88.1\%$ across the five alignment methods,
but reconstruction performance does not consistently follow this
ordering. SAMGA achieves the highest retrieval accuracy, whereas
Brain-HiVE outperforms it on six of the seven reconstruction metrics.

The no-alignment control remains competitive in low-level fidelity,
ranking second in PixCorr and third in SSIM. However, all five
pretrained representations outperform it on AlexNet(2), AlexNet(5),
and Inception, and four also achieve better CLIP and SwAV scores.
Thus, alignment pretraining benefits high-level reconstruction in
these experiments, but stronger retrieval performance does not
necessarily translate into better reconstruction.

This comparison is descriptive, as the alignment methods differ
in their pretraining objectives and representation designs.

\begin{figure*}[t]
  \centering
  \includegraphics[width=1\linewidth]{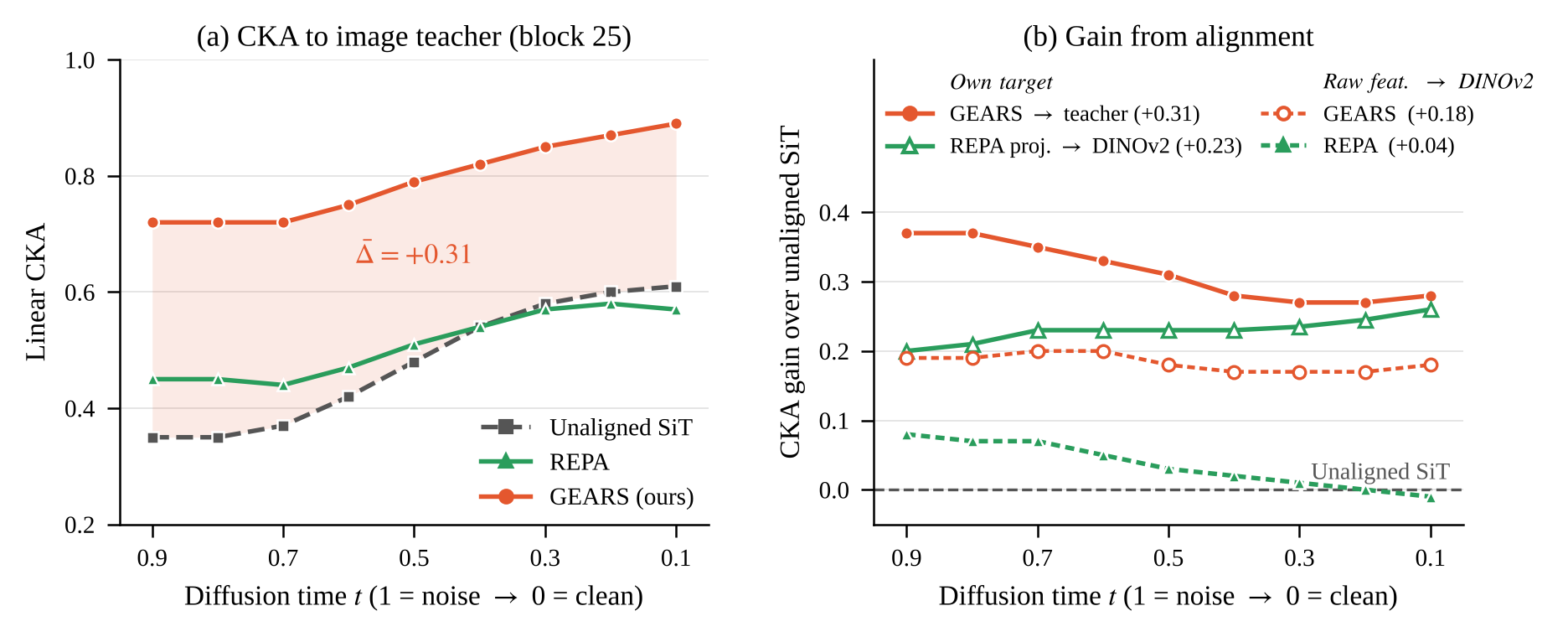}
  \caption{
Baseline-controlled CKA analysis averaged over 10 THINGS-EEG subjects.
(a) CKA between student block-8 features and image-teacher block-25 features.
(b) CKA gains over unaligned SiT: solid lines compare each model with its training target (REPA uses projector outputs), while dashed lines compare raw block-8 features with DINOv2.
$\bar{\Delta}$ denotes the mean gain across timesteps.  
  }
  \label{fig:cka_baseline}
\end{figure*}

\section{Analysis of Alignment Targets}
\label{app:cka_baseline}

\subsection{Linear CKA}
We measure representational similarity with linear centered kernel alignment (CKA)~\cite{kornblith2019similarity}. Given feature matrices $\mathbf{X} \in \mathbb{R}^{n \times d_1}$ and $\mathbf{Y} \in \mathbb{R}^{n \times d_2}$ computed on the same $n$ samples and centered over samples,
\begin{equation}
\mathrm{CKA}(\mathbf{X}, \mathbf{Y}) =
\frac{\left\| \mathbf{Y}^{\top} \mathbf{X} \right\|_F^2}
{\left\| \mathbf{X}^{\top} \mathbf{X} \right\|_F \, \left\| \mathbf{Y}^{\top} \mathbf{Y} \right\|_F}.
\end{equation}
CKA ranges from 0 to 1 and is invariant to orthogonal transformations and isotropic scaling, allowing comparison between representations of different dimensionality.

\subsection{Baseline-Controlled Representation Similarity Analysis}
\paragraph{Motivation}
In the main paper (Fig.~2c and Fig.~6), we compare the absolute CKA of each student to its own alignment target. Here, we additionally measure the unaligned SiT-XL/2 as a reference and compare the relative gain of each alignment over this baseline.

\paragraph{Setup}
We compare three EEG-conditioned students: the unaligned SiT-XL/2 (Table~1, SiT-XL/2$^\dagger$), REPA\cite{yurepresentation} aligned to DINOv2, and GEARS aligned to the image teacher. For each model, we compute CKA between its block-8 features and (i) the teacher's block-25 features and (ii) DINOv2 features, using the same test images, noise seeds, and timesteps, and average over all 10 subjects. All models are trained for 200 epochs. Since REPA applies its loss through an MLP projector~\cite{yurepresentation}, we report both its raw block-8 features and its projector output; GEARS uses no projector. We denote by $\bar{\Delta}$ the CKA increase over the unaligned model, averaged over timesteps.

\paragraph{Results}
As shown in Fig.~\ref{fig:cka_baseline} (a), the unaligned model already shows moderate CKA to the teacher (0.35--0.61), which rises toward $t \to 0$. Since no alignment is involved, this similarity reflects the shared noisy input, and part of the gap in Fig.~2c therefore comes from this baseline.

Fig.~\ref{fig:cka_baseline} (b) shows the gain of each alignment over this baseline. When each model is compared with its own alignment target (solid lines), GEARS increases CKA to the teacher by $+0.31$, exceeding the $+0.23$ gain of REPA even when REPA is evaluated in its projector output, where its loss is applied. However, this gain of REPA is largely confined to the projector: its raw block-8 features increase CKA to DINOv2 by only $+0.04$ (dashed lines). This suggests that the projector bridges most of the gap between the EEG-conditioned features and DINOv2, so that the student itself adopts little of the visual structure of DINOv2. In contrast, GEARS aligns the raw features directly without a projector, and even increases their CKA to DINOv2 by $+0.18$ without being aligned to it. These results show that the teacher target is followed more closely even after controlling for the baseline, and that its effect reaches the backbone representations used for generation.

\begin{figure}
    \centering
    \includegraphics[width=1\linewidth]{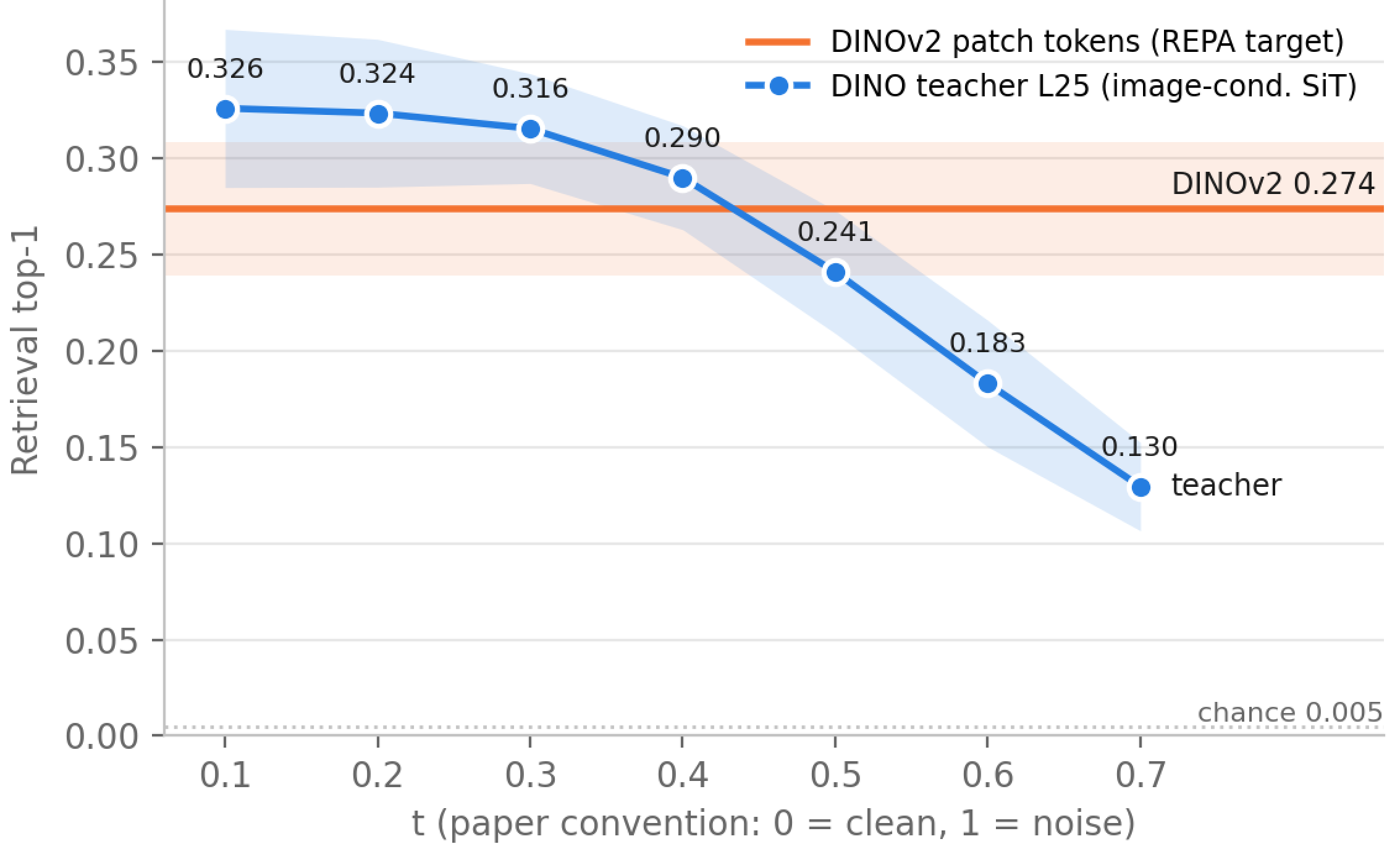}
    \caption{Predictability of alignment targets from EEG on THINGS-EEG. We predict each target from the EEG condition with ridge regression and report 200-way retrieval top-1 accuracy (mean $\pm$ SD over 10 subjects). The teacher block-25 features are more predictable than DINOv2 patch tokens at low noise levels.}
    \label{fig:predictability}
\end{figure}

\subsection{Predictability of Alignment Targets from EEG}
\label{app:predictability}

The main paper shows that the EEG-conditioned student follows the teacher features more closely than DINOv2 features. However, the student features used in that CKA analysis are computed from both the EEG condition and the noisy latent $\mathbf{z}_t$. To separate the contribution of EEG, we predict each target solely from the EEG condition. For each subject, we fit a ridge regression from the EEG condition to each target on the training set and evaluate 200-way retrieval on the test set. We use retrieval as the evaluation metric because it directly measures whether the prediction carries image-specific information that identifies the correct image among all test images.

As shown in Fig.~\ref{fig:predictability}, the teacher features are easier to predict from EEG than DINOv2 when the noise level is low ($t \le 0.4$). This suggests that the teacher expresses the same visual information in a form that EEG can more easily match. When the noise level is high ($t \ge 0.5$), the teacher features become harder to predict, because they increasingly depend on the random noise in $\mathbf{z}_t$, which EEG does not contain. However, the student also receives $\mathbf{z}_t$, so it can still access this noise-dependent part during training. This analysis therefore underestimates how well the student can follow the teacher target.

\begin{table}[t]
\centering
\caption{Reconstruction performance of the image-conditioned teacher on the THINGS-EEG test set, compared with EEG-conditioned models.}
\resizebox{\columnwidth}{!}{%
\begin{tabular}{lccccccc}
\toprule
Method & PixCorr$\uparrow$ & SSIM$\uparrow$ & Alex(2)$\uparrow$ & Alex(5)$\uparrow$ & Incep.$\uparrow$ & CLIP$\uparrow$ & SwAV$\downarrow$ \\
\midrule
SiT-XL/2 & 0.261 & 0.346 & 0.916 & 0.941 & 0.799 & 0.791 & 0.508 \\
Ours    & \textbf{0.263} & 0.370 & 0.921 & 0.942 & 0.813 & 0.799 & 0.500 \\
\midrule
Teacher & 0.238 & \textbf{0.381} & \textbf{0.944} & \textbf{0.985} & \textbf{0.990} & \textbf{0.960} & \textbf{0.331} \\
\bottomrule
\end{tabular}}
\label{tab:teacher}
\end{table}

\begin{table*}[t]
\centering
\caption{Full results of the alignment-target ablation on THINGS-EEG (mean over 10 subjects). All models share the same EEG-conditioned SiT backbone and differ only in the alignment target. Bold and underline denote the best and second-best results.}

\setlength{\tabcolsep}{5pt}
\begin{tabular}{lccccccc}
\toprule
\multirow{2}{*}{Alignment target}
& \multicolumn{2}{c}{Low-level} & \multicolumn{5}{c}{High-level} \\
\cmidrule(lr){2-3} \cmidrule(lr){4-8}
& PixCorr$\uparrow$ & SSIM$\uparrow$ & AlexNet(2)$\uparrow$ & AlexNet(5)$\uparrow$ & Inception$\uparrow$ & CLIP$\uparrow$ & SwAV$\downarrow$ \\
\midrule
None (SiT)          & 0.261 & 0.346 & 0.916 & 0.941 & 0.799 & 0.791 & 0.508 \\
\midrule
DINOv2              & 0.260 & 0.349 & 0.913 & 0.940 & \underline{0.807} & \underline{0.797} & \underline{0.502} \\
VAE                 & \textbf{0.264} & \underline{0.360} & \underline{0.918} & \textbf{0.942} & \textbf{0.813} & 0.789 & 0.506 \\
Teacher, null cond. & 0.244 & 0.357 & 0.903 & 0.929 & 0.790 & 0.777 & 0.513 \\
\rowcolor{gray!12}
\textbf{Teacher (ours)} & \underline{0.263} & \textbf{0.370} & \textbf{0.921} & \textbf{0.942} & \textbf{0.813} & \textbf{0.799} & \textbf{0.500} \\
\bottomrule
\end{tabular}
\label{tab:align_target_full}
\end{table*}

\begin{table*}[t]
\centering
\caption{Comparison of teacher transfer strategies on THINGS-EEG, averaged over 10 subjects. All methods use the same frozen image-conditioned teacher and differ only in where its knowledge is transferred. Bold denotes the best result.}
\setlength{\tabcolsep}{4pt}
\begin{tabular}{llccccccc}
\toprule
\multirow{2}{*}{Method} & \multirow{2}{*}{Supervises}
& \multicolumn{2}{c}{Low-level} & \multicolumn{5}{c}{High-level} \\
\cmidrule(lr){3-4} \cmidrule(lr){5-9}
& & PixCorr$\uparrow$ & SSIM$\uparrow$ & AlexNet(2)$\uparrow$ & AlexNet(5)$\uparrow$ & Inception$\uparrow$ & CLIP$\uparrow$ & SwAV$\downarrow$ \\
\midrule
Condition alignment   & Input    & 0.228 & 0.365 & 0.911 & 0.934 & 0.796 & 0.782 & 0.513 \\
Velocity distillation & Output   & \textbf{0.264} & 0.353 & 0.920 & \textbf{0.943} & 0.811 & 0.790 & 0.506 \\
\textbf{GEARS (ours)} & Features & 0.263 & \textbf{0.370} & \textbf{0.921} & 0.942 & \textbf{0.813} & \textbf{0.799} & \textbf{0.500} \\
\bottomrule
\end{tabular}
\label{tab:transfer_full}
\end{table*}

\section{Image-Conditioned Teacher}
\label{app:teacher}

\subsection{Reconstruction Performance}
The image-conditioned teacher is trained once on the THINGS-EEG training images only, without any EEG input, and is shared across all subjects. Table~\ref{tab:teacher} reports its reconstruction performance on the test images. Conditioned on DINOv2 features of the target image, the teacher achieves near-ceiling high-level scores and the highest SSIM, while its PixCorr is lower than that of the EEG-conditioned models. Since the teacher receives a single global DINOv2 embedding through AdaLN, it preserves semantics and coarse structure but not precise pixel-level layout, whereas the Brain-HiVE EEG embedding is partly trained with VAE features. This indicates that the teacher does not act as a pixel-level reconstruction prior; the gains of GEARS appear in SSIM and high-level metrics, where the teacher is stronger than the EEG-conditioned student.

\subsection{Layer-wise Features}
\begin{figure}[t]
  \centering
  \includegraphics[width=\linewidth]{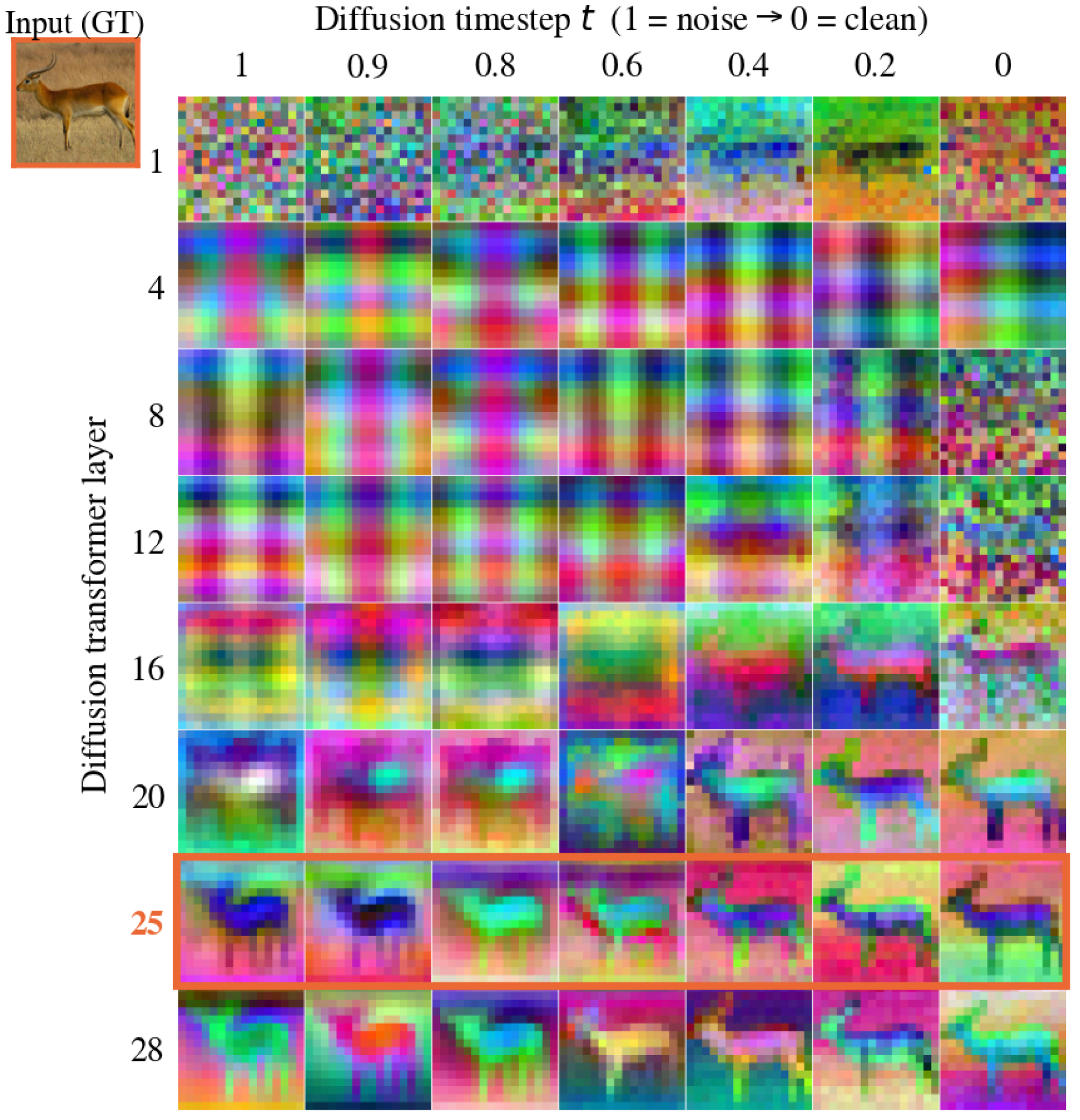}
  \caption{PCA visualization of image-teacher features across layers and timesteps for a test image. Deep layers reveal the object's layout even near pure noise. Layer 25 (orange) is used as the alignment target.}
  \label{fig:teacher_pca}
\end{figure}

To choose which teacher layer to use as the alignment target, we visualize the teacher's intermediate features across layers and timesteps (Fig.~\ref{fig:teacher_pca}). We project the patch tokens of each layer onto their top-3 principal components and map them to RGB; PCA is fitted separately for each layer and timestep, so colors are not comparable across cells. Shallow layers are dominated by positional patterns and show little image-specific structure. In contrast, deeper layers, from around layer 20 onward, reveal the object's shape and layout, which become clearer as $t \to 0$. Notably, layer 25 already shows a coarse object layout at $t=1$, where the input latent is pure noise, indicating that this structure is derived from the DINOv2 condition rather than from the noisy input. Based on this observation, we use a deep teacher layer as the alignment target, and select layer 25 among the deep layers on the validation split (Appendix~\ref{app:hparam}).

\begin{table*}[t]
\centering
\caption{Full results of the generalization experiment on THINGS-EEG under three EEG conditioning representations (mean over 10 subjects; all generators trained for 100 epochs). SiT denotes the same backbone trained without alignment.}
\setlength{\tabcolsep}{5pt}
\begin{tabular}{llccccccc}
\toprule
\multirow{2}{*}{Condition} & \multirow{2}{*}{Generator}
& \multicolumn{2}{c}{Low-level} & \multicolumn{5}{c}{High-level} \\
\cmidrule(lr){3-4} \cmidrule(lr){5-9}
& & PixCorr$\uparrow$ & SSIM$\uparrow$ & AlexNet(2)$\uparrow$ & AlexNet(5)$\uparrow$ & Inception$\uparrow$ & CLIP$\uparrow$ & SwAV$\downarrow$ \\
\midrule
\multirow{4}{*}{UBP}
& IP-Adapter & 0.184 & \textbf{0.379} & 0.810 & 0.859 & 0.717 & 0.742 & 0.601 \\
& AVDE       & 0.160 & 0.369 & 0.823 & 0.888 & 0.760 & 0.742 & 0.545 \\
& SiT        & 0.236 & 0.330 & 0.893 & 0.926 & 0.787 & 0.757 & 0.547 \\
& \cellcolor{gray!12}\textbf{Ours}
& \cellcolor{gray!12}\textbf{0.245} & \cellcolor{gray!12}0.363 & \cellcolor{gray!12}\textbf{0.907} & \cellcolor{gray!12}\textbf{0.937} & \cellcolor{gray!12}\textbf{0.801} & \cellcolor{gray!12}\textbf{0.802} & \cellcolor{gray!12}\textbf{0.508} \\
\midrule
\multirow{4}{*}{Brain-HiVE}
& IP-Adapter & 0.188 & 0.355 & 0.790 & 0.837 & 0.710 & 0.724 & 0.603 \\
& AVDE       & 0.164 & 0.365 & 0.811 & 0.868 & 0.743 & 0.736 & 0.546 \\
& SiT        & 0.261 & 0.360 & \textbf{0.910} & \textbf{0.935} & \textbf{0.803} & 0.787 & 0.516 \\
& \cellcolor{gray!12}\textbf{Ours}
& \cellcolor{gray!12}\textbf{0.263} & \cellcolor{gray!12}\textbf{0.377} & \cellcolor{gray!12}0.908 & \cellcolor{gray!12}0.932 & \cellcolor{gray!12}0.798 & \cellcolor{gray!12}\textbf{0.792} & \cellcolor{gray!12}\textbf{0.513} \\
\midrule
\multirow{4}{*}{SAMGA}
& IP-Adapter & 0.193 & 0.351 & 0.807 & 0.868 & 0.718 & 0.744 & 0.582 \\
& AVDE       & 0.161 & 0.367 & 0.834 & 0.887 & 0.755 & 0.730 & 0.537 \\
& SiT        & 0.217 & 0.357 & 0.893 & 0.924 & 0.790 & 0.789 & 0.522 \\
& \cellcolor{gray!12}\textbf{Ours}
& \cellcolor{gray!12}\textbf{0.257} & \cellcolor{gray!12}\textbf{0.384} & \cellcolor{gray!12}\textbf{0.912} & \cellcolor{gray!12}\textbf{0.933} & \cellcolor{gray!12}\textbf{0.803} & \cellcolor{gray!12}\textbf{0.798} & \cellcolor{gray!12}\textbf{0.511} \\
\bottomrule
\end{tabular}
\label{tab:generalization_full}
\end{table*}

\section{Additional Ablation Results}
\subsection{Comparison of Supervision Targets}
\label{app:align_target}

Table~\ref{tab:align_target_full} reports the full results of the alignment-target ablation in Fig.~5 of the main paper on all seven metrics. Fig.~5 summarizes these results as follows: for each metric, we compute the relative gain over the unaligned SiT, $(x - x_{\mathrm{SiT}}) / x_{\mathrm{SiT}} \times 100$, using the 10-subject means, with the sign flipped for SwAV so that a positive value indicates improvement. The low-level gain is the average over PixCorr and SSIM, and the high-level gain is the average over CLIP and SwAV.

The per-metric results are consistent with this summary. Aligning to DINOv2 improves CLIP and SwAV but leaves PixCorr and SSIM nearly unchanged, whereas aligning to VAE improves PixCorr and SSIM but not CLIP. Aligning to the teacher under the null condition lowers all metrics except SSIM below the unaligned SiT. In contrast, aligning to the image-conditioned teacher achieves the best or tied-best result on six of the seven metrics and is within 0.001 of the best on PixCorr.

\subsection{Comparison of Teacher Transfer Strategies}
\label{app:transfer}

To identify where teacher knowledge should be transferred, we compare GEARS with two alternatives that use the same frozen teacher at different stages of the generator: condition alignment at the input and velocity distillation at the output.

\paragraph{Condition alignment.}
This baseline applies a contrastive objective between the projected EEG condition $\mathbf{c}_e$ and the image condition $\mathbf{c}_v$ before they are injected into the generator:
\begin{equation}
\mathcal{L}_{\mathrm{cond}}
= -\frac{1}{B}\sum_{i=1}^{B}
\log
\frac{\exp\!\left(\operatorname{sim}\!\left(\mathbf{c}_{e}^{(i)}, \operatorname{sg}[\mathbf{c}_{v}^{(i)}]\right)/\tau\right)}
{\sum_{j=1}^{B}\exp\!\left(\operatorname{sim}\!\left(\mathbf{c}_{e}^{(i)}, \operatorname{sg}[\mathbf{c}_{v}^{(j)}]\right)/\tau\right)},
\end{equation}
where $B$ is the batch size, $\operatorname{sim}(\cdot,\cdot)$ is cosine similarity, $\tau$ is a temperature, and $\operatorname{sg}[\cdot]$ is the stop-gradient. It makes the two conditions more similar but does not constrain how the generator processes them.

\paragraph{Velocity distillation.}
Following output-level diffusion distillation~\cite{salimans2022progressive}, we feed the same noisy latent $\mathbf{z}_t$ and timestep $t$ to the frozen teacher and the student, and match their velocity predictions:
\begin{equation}
\mathcal{L}_{\mathrm{vel}}
= \left\| \mathbf{v}^S_{\phi}(\mathbf{z}_t, t, \mathbf{c}_e)
- \operatorname{sg}\!\left[\mathbf{v}^T_{\theta}(\mathbf{z}_t, t, \mathbf{c}_v)\right] \right\|_2^2.
\end{equation}

Each transfer loss replaces $\mathcal{L}_{\mathrm{align}}$ in Eq.~(10) of the main paper, i.e., $\mathcal{L} = \mathcal{L}_{\mathrm{velocity}} + \lambda\,\mathcal{L}_{\mathrm{transfer}}$ with $\mathcal{L}_{\mathrm{transfer}} \in \{\mathcal{L}_{\mathrm{cond}}, \mathcal{L}_{\mathrm{vel}}\}$. All other settings follow the main experiments.

\paragraph{Results.}
As shown in Table~\ref{tab:transfer_full}, condition alignment performs worst on most metrics, with the largest drop in PixCorr, indicating that making the EEG condition resemble the image condition does not by itself improve reconstruction. Velocity distillation is more competitive and is within 0.001 of GEARS on PixCorr and AlexNet(5), but GEARS performs better on the remaining five metrics, with the largest margins in SSIM (0.353 $\rightarrow$ 0.370) and CLIP (0.790 $\rightarrow$ 0.799). Since condition alignment acts only on the input of the generator and velocity distillation only on its output, these results suggest that the teacher is most effective when it supervises the intermediate representations of the generator.

\begin{figure*}[t]
  \centering
  \includegraphics[width=\linewidth]{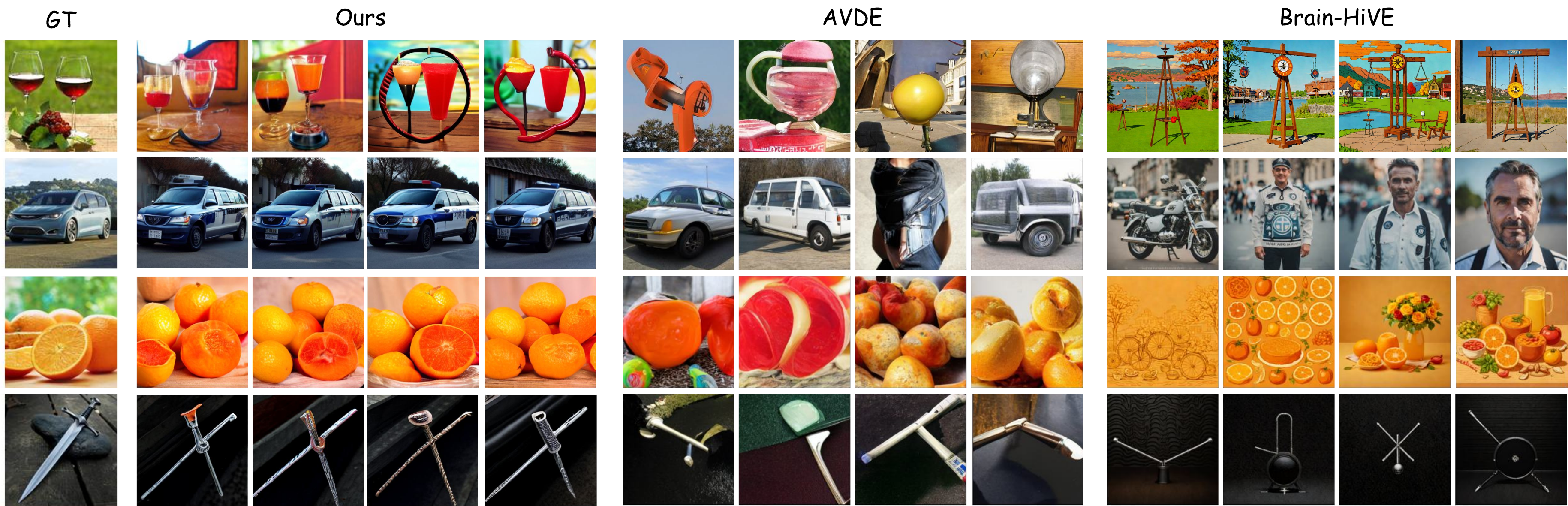}
  \caption{Qualitative comparison of seed robustness on THINGS-EEG. For each stimulus (left), we show reconstructions from models trained with four different seeds for our method, AVDE, and IP-Adapter (Brain-HiVE), from left to right.}
  \label{fig:seed_sample}
\end{figure*}

\begin{figure*}[t]
  \centering
  \includegraphics[width=0.85\linewidth]{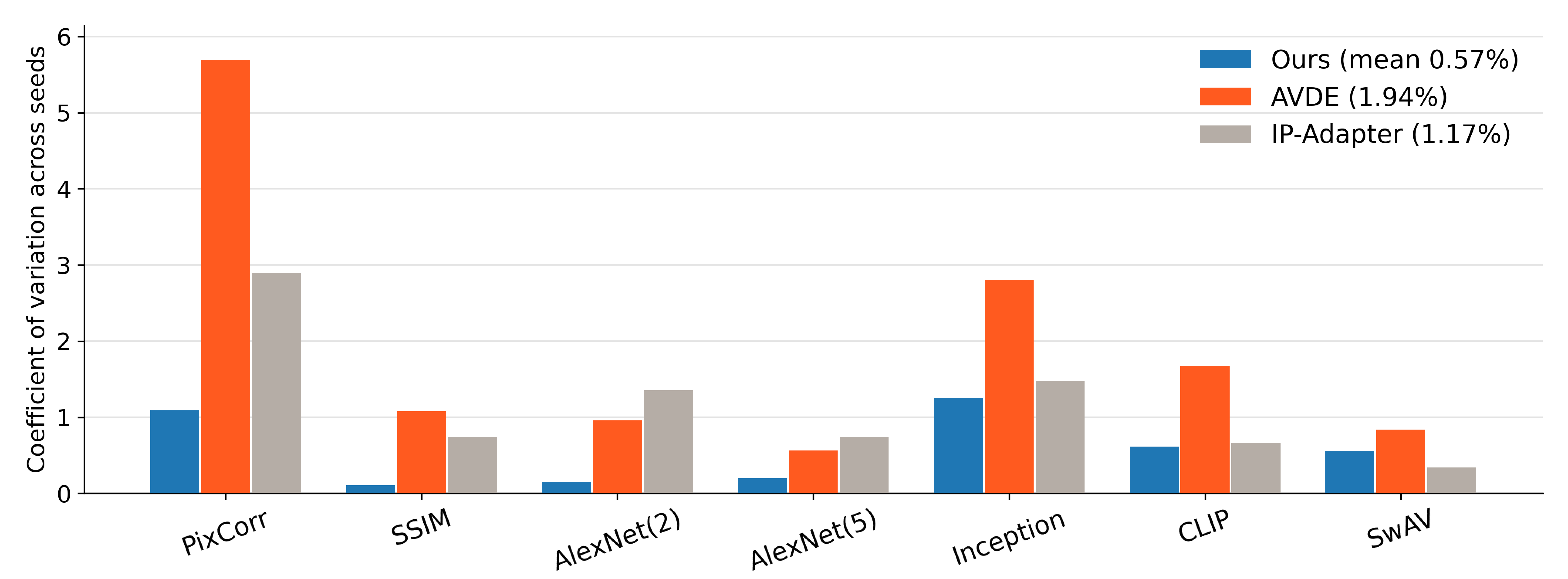}
    \caption{Coefficient of variation (CV) of each metric across 10 training seeds on THINGS-EEG (subject~8). Lower is more stable. Average CV over the seven metrics is shown in the legend.}

  \label{fig:seed_robustness}
\end{figure*}

\section{Further Analysis}

\subsection{Computational Cost}
\label{app:cost}

Table~\ref{tab:cost} compares the training time of each method, separating one-time pretraining from subject-specific EEG training. Our image-conditioned teacher is trained once and shared across all subjects, requiring less time than the IP-Adapter pretraining of Brain-HiVE. For subject-specific training, Brain-HiVE is the most efficient because it keeps the IP-Adapter and image generator frozen and optimizes only the EEG encoder. Our method fully fine-tunes the generator and is therefore more costly, but requires less time than AVDE, which also fully fine-tunes its generator. Since the teacher is discarded after training, our method adds no cost at inference.

\begin{table}[h]
\centering
\caption{Training time on a single NVIDIA B200 GPU. Pretraining denotes the one-time stage before subject-specific training: IP-Adapter pretraining for Brain-HiVE and teacher training for ours. EEG training is measured per subject for 200 epochs.}
\setlength{\tabcolsep}{8pt}
\begin{tabular}{lcc}
\toprule
Method & Pretraining & EEG training \\
\midrule
AVDE       & --     & 2h 49m \\
Brain-HiVE & 6h 15m & 5m 12s \\
Ours       & 2h 07m & 2h 28m \\
\bottomrule
\end{tabular}
\label{tab:cost}
\end{table}

\subsection{Robustness Across Training Seeds}
\label{app:seed}

\paragraph{Why Seed Robustness Matters for Reconstruction}
Unlike unconditional or class-conditional generation, EEG-to-image reconstruction aims to recover a specific visual stimulus from the corresponding neural signal. A reliable reconstruction model should therefore achieve consistent performance across training seeds. High seed-to-seed variation would indicate that reconstruction quality depends on random initialization and stochastic optimization rather than on a stable mapping from EEG to the target image, reducing the reproducibility of the decoded result.

\paragraph{Evaluation Protocol}
We compare our method with two recent EEG-to-image reconstruction models based on distinct generative paradigms: AVDE, which uses a visual autoregressive model, and Brain-HiVE, which uses SDXL with an IP-Adapter. For a fair comparison, all methods are conditioned on the EEG representation of Brain-HiVE. Each method is trained with $K=10$ different seeds on subject 8, and for each metric $m$ we compute the coefficient of variation (CV) across seeds:
\begin{equation}
\mathrm{CV}_{m} = 100 \cdot \frac{\operatorname{Std}_{s=1}^{K}\left(q_{m,s}\right)}{\left|\operatorname{Mean}_{s=1}^{K}\left(q_{m,s}\right)\right|},
\end{equation}
where $q_{m,s}$ is the performance on metric $m$ of the model trained with seed $s$. A lower CV indicates lower sensitivity to the training seed. We report the CV for each metric and its average over the seven metrics.

\paragraph{Results}
As shown in Fig.~\ref{fig:seed_robustness}, our method attains the lowest average CV (0.57\%), compared with 1.94\% for AVDE and 1.17\% for the IP-Adapter pipeline. It achieves the lowest CV on six of the seven metrics; the IP-Adapter pipeline is lower only on SwAV. The reduction is most pronounced on the low-level metrics, particularly PixCorr, where the other methods vary the most. Fig.~\ref{fig:seed_sample} shows that our reconstructions consistently preserve the category, layout, and color of the stimulus across seeds, whereas those of the other methods vary in object identity and composition.

\begin{table}[t]
\centering
\caption{Paired comparison between our method and the unaligned SiT-XL/2 on THINGS-EEG. Bold denotes $p<0.05$; $p$-values are uncorrected for multiple comparisons.}
\resizebox{\linewidth}{!}{%
\begin{tabular}{lccc}
\toprule
Metric & Mean $\Delta$ & Subjects improved & $p$ \\
\midrule
PixCorr        & $+0.002$ & 6/10  & 0.695 \\
SSIM           & $+0.024$ & 10/10 & \textbf{0.002} \\
AlexNet(2)     & $+0.005$ & 8/10  & \textbf{0.020} \\
AlexNet(5)     & $+0.001$ & 5/10  & 0.492 \\
Inception      & $+0.014$ & 9/10  & \textbf{0.020} \\
CLIP           & $+0.008$ & 8/10  & \textbf{0.010} \\
SwAV & $+0.008$ & 10/10 & \textbf{0.002} \\
\bottomrule
\end{tabular}}

\label{tab:stats}
\end{table}

\begin{table*}[t]
\centering
\caption{Subject-wise reconstruction performance of our method on THINGS-MEG. Each subject's result is averaged over five sampling seeds.}
\setlength{\tabcolsep}{4pt}
\begin{tabular}{lccccccc}
\toprule
\multirow{2}{*}{Subject}
& \multicolumn{2}{c}{Low-level} & \multicolumn{5}{c}{High-level} \\
\cmidrule(lr){2-3} \cmidrule(lr){4-8}
& PixCorr$\uparrow$ & SSIM$\uparrow$ & AlexNet(2)$\uparrow$ & AlexNet(5)$\uparrow$ & Inception$\uparrow$ & CLIP$\uparrow$ & SwAV$\downarrow$ \\
\midrule
1 & 0.112 & 0.276 & 0.694 & 0.816 & 0.692 & 0.705 & 0.587 \\
2 & 0.235 & 0.348 & 0.883 & 0.933 & 0.823 & 0.794 & 0.520 \\
3 & 0.167 & 0.348 & 0.856 & 0.920 & 0.803 & 0.780 & 0.536 \\
4 & 0.135 & 0.341 & 0.766 & 0.825 & 0.728 & 0.701 & 0.585 \\
\midrule
\textbf{Mean} & \textbf{0.162} & \textbf{0.328} & \textbf{0.800} & \textbf{0.874} & \textbf{0.762} & \textbf{0.745} & \textbf{0.557} \\
\bottomrule
\end{tabular}
\label{tab:meg_subject_results}
\end{table*}

\begin{table*}[t]
\centering
\caption{Subject-wise reconstruction performance of our method on THINGS-EEG. Each subject's result is averaged over five sampling seeds.}
\setlength{\tabcolsep}{4pt}
\begin{tabular}{lccccccc}
\toprule
\multirow{2}{*}{Subject}
& \multicolumn{2}{c}{Low-level} & \multicolumn{5}{c}{High-level} \\
\cmidrule(lr){2-3} \cmidrule(lr){4-8}
& PixCorr$\uparrow$ & SSIM$\uparrow$ & AlexNet(2)$\uparrow$ & AlexNet(5)$\uparrow$ & Inception$\uparrow$ & CLIP$\uparrow$ & SwAV$\downarrow$ \\
\midrule
1  & 0.252 & 0.384 & 0.922 & 0.939 & 0.761 & 0.741 & 0.536 \\
2  & 0.264 & 0.378 & 0.932 & 0.942 & 0.816 & 0.804 & 0.497 \\
3  & 0.284 & 0.357 & 0.922 & 0.939 & 0.829 & 0.801 & 0.494 \\
4  & 0.236 & 0.376 & 0.910 & 0.942 & 0.832 & 0.824 & 0.497 \\
5  & 0.248 & 0.328 & 0.894 & 0.924 & 0.767 & 0.776 & 0.525 \\
6  & 0.270 & 0.353 & 0.925 & 0.944 & 0.804 & 0.797 & 0.499 \\
7  & 0.222 & 0.369 & 0.913 & 0.931 & 0.799 & 0.784 & 0.509 \\
8  & 0.309 & 0.397 & 0.947 & 0.961 & 0.857 & 0.824 & 0.478 \\
9  & 0.289 & 0.368 & 0.907 & 0.938 & 0.803 & 0.790 & 0.499 \\
10 & 0.261 & 0.393 & 0.942 & 0.964 & 0.857 & 0.846 & 0.470 \\
\midrule
\textbf{Mean} & \textbf{0.263} & \textbf{0.370} & \textbf{0.921} & \textbf{0.942} & \textbf{0.813} & \textbf{0.799} & \textbf{0.500} \\
\bottomrule
\end{tabular}
\label{tab:eeg_subject_results}
\end{table*}

\section{Additional Quantitative Results}

\subsection{Generalization Across EEG Conditions}
\label{app:generalization}

Table~\ref{tab:generalization_full} reports the full results of the generalization experiment in Table~4 of the main paper on all seven metrics. Under SAMGA, GEARS achieves the best result on all seven metrics. Under UBP, it achieves the best result on all metrics except SSIM, where the IP-Adapter pipeline is higher. Under Brain-HiVE, GEARS achieves the best low-level metrics, CLIP, and SwAV, while the unaligned SiT is slightly higher on AlexNet and Inception, by at most 0.005. Overall, GEARS improves over the unaligned SiT more under the weaker conditions, UBP and SAMGA, consistent with the main paper.

\subsection{Subject-wise Results}
\label{app:subject}
Tables~\ref{tab:eeg_subject_results} and~\ref{tab:meg_subject_results} report the subject-wise reconstruction performance of our method on THINGS-EEG and THINGS-MEG, respectively, corresponding to the mean values in Table~1 of the main paper. Each subject's result is averaged over five sampling seeds. Performance varies across subjects, particularly on THINGS-MEG, where subjects 1 and 4 show lower scores than subjects 2 and 3 on all metrics.

\subsection{Statistical Comparison with the Unaligned Baseline}
\label{app:stats}

Table~\ref{tab:stats} reports the paired comparison between our method and the unaligned SiT-XL/2 on THINGS-EEG, which share the same backbone and EEG condition. For each metric, we report the mean difference, the number of subjects for which our method improves, and the $p$-value of a two-sided paired Wilcoxon signed-rank test across subjects. Our method improves SSIM and SwAV for all 10 subjects, and the improvements in SSIM, AlexNet(2), Inception, CLIP, and SwAV are significant. We correct an omission in the main paper: the improvement in AlexNet(2) is also statistically significant ($p=0.020$),
bringing the total to five metrics under the uncorrected
two-sided paired Wilcoxon signed-rank test.

\begin{figure*}[p]
    \centering
    \includegraphics[width=1\linewidth]{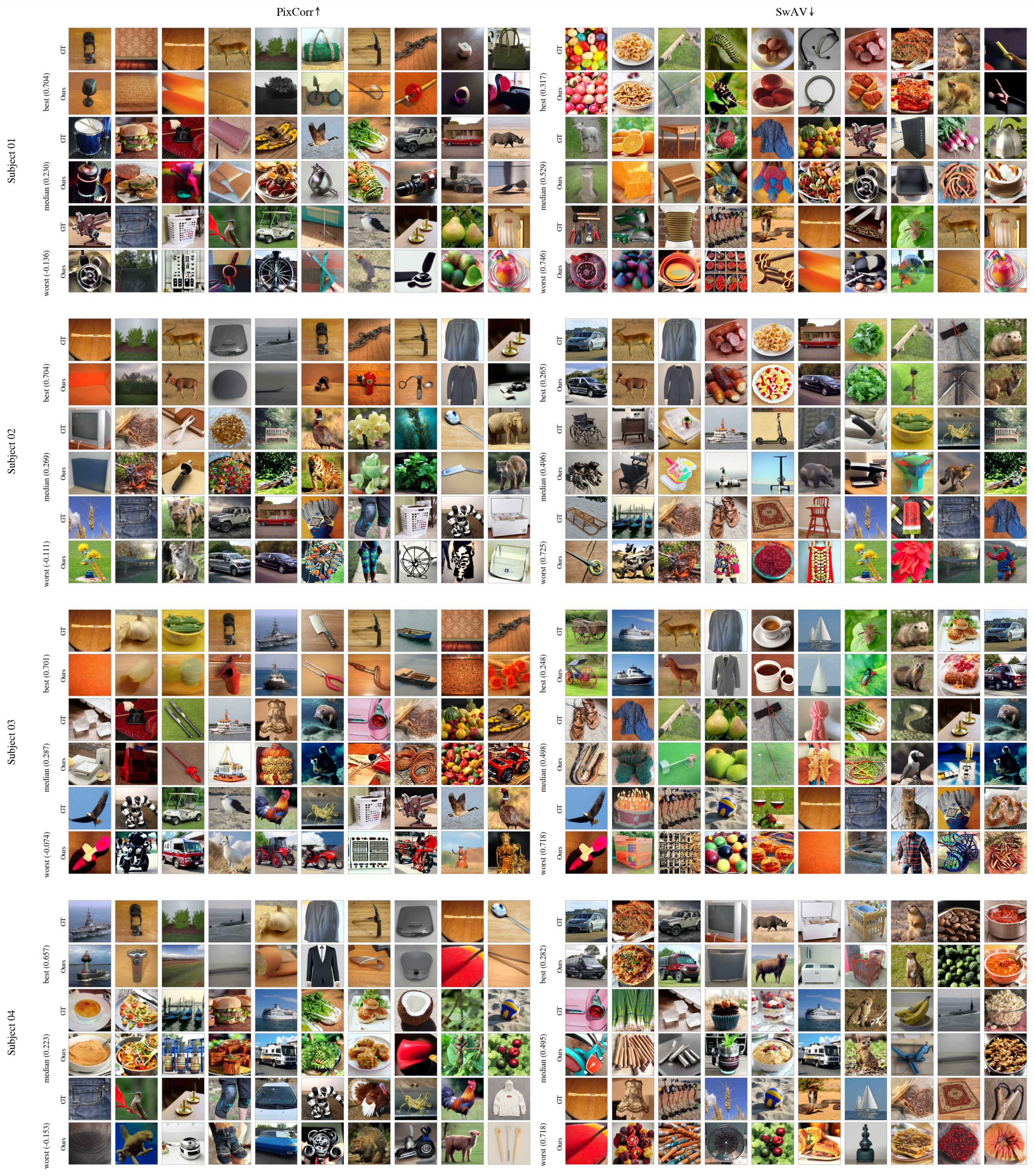}
    \caption{Additional qualitative results for subjects 1-4. For both PixCorr and SwAV, we show the best, median, and worst reconstruction samples selected
    independently from the test set.}
    \label{fig:qr1}
\end{figure*}

\begin{figure*}[p]
    \centering
    \includegraphics[width=1\linewidth]{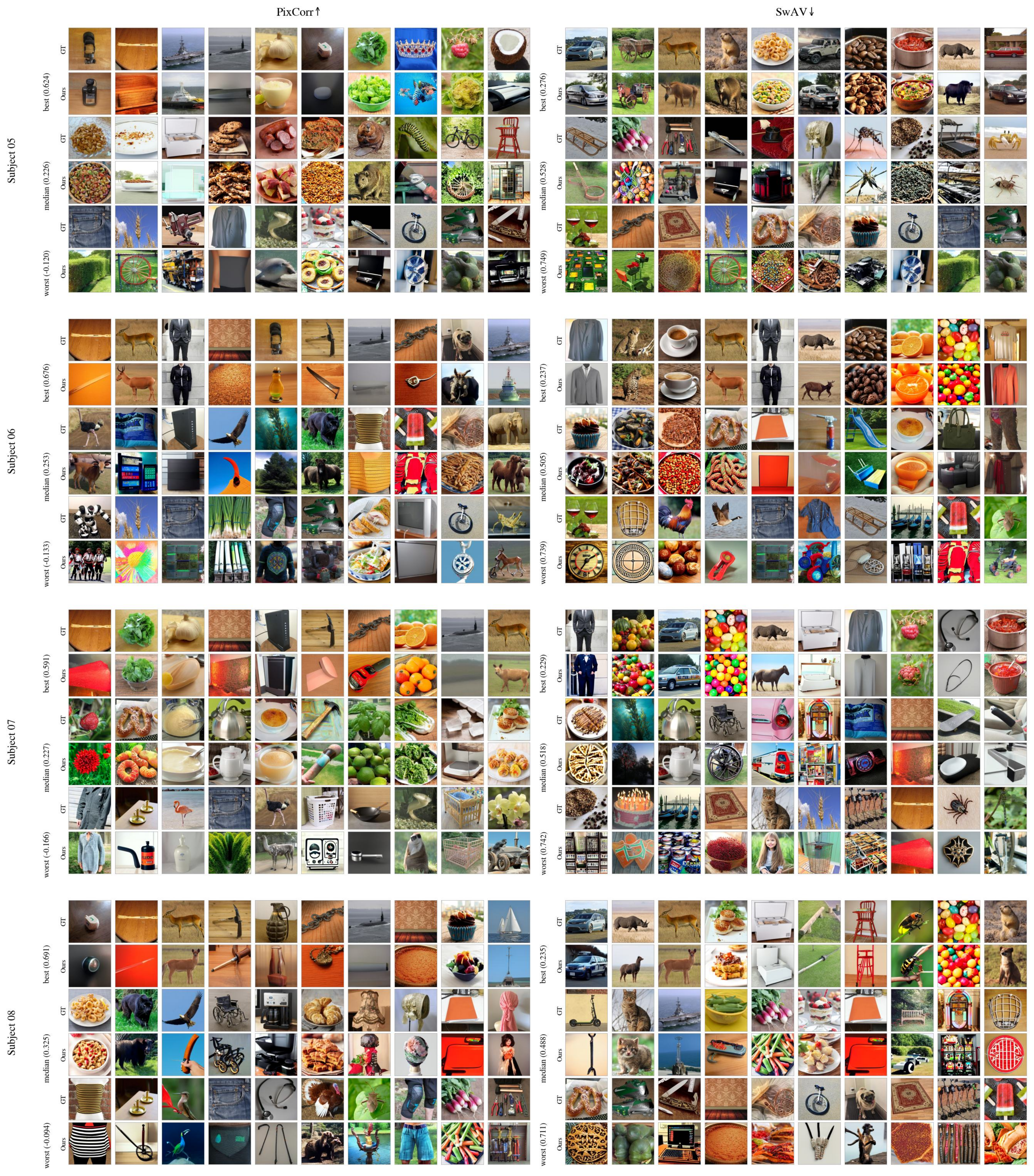}
    \caption{Additional qualitative results for subjects 5-8. For both PixCorr and SwAV, we show the best, median, and worst reconstruction samples selected
    independently from the test set.}
    \label{fig:qr1}
\end{figure*}

\begin{figure*}[p]
    \centering
    \includegraphics[width=1\linewidth]{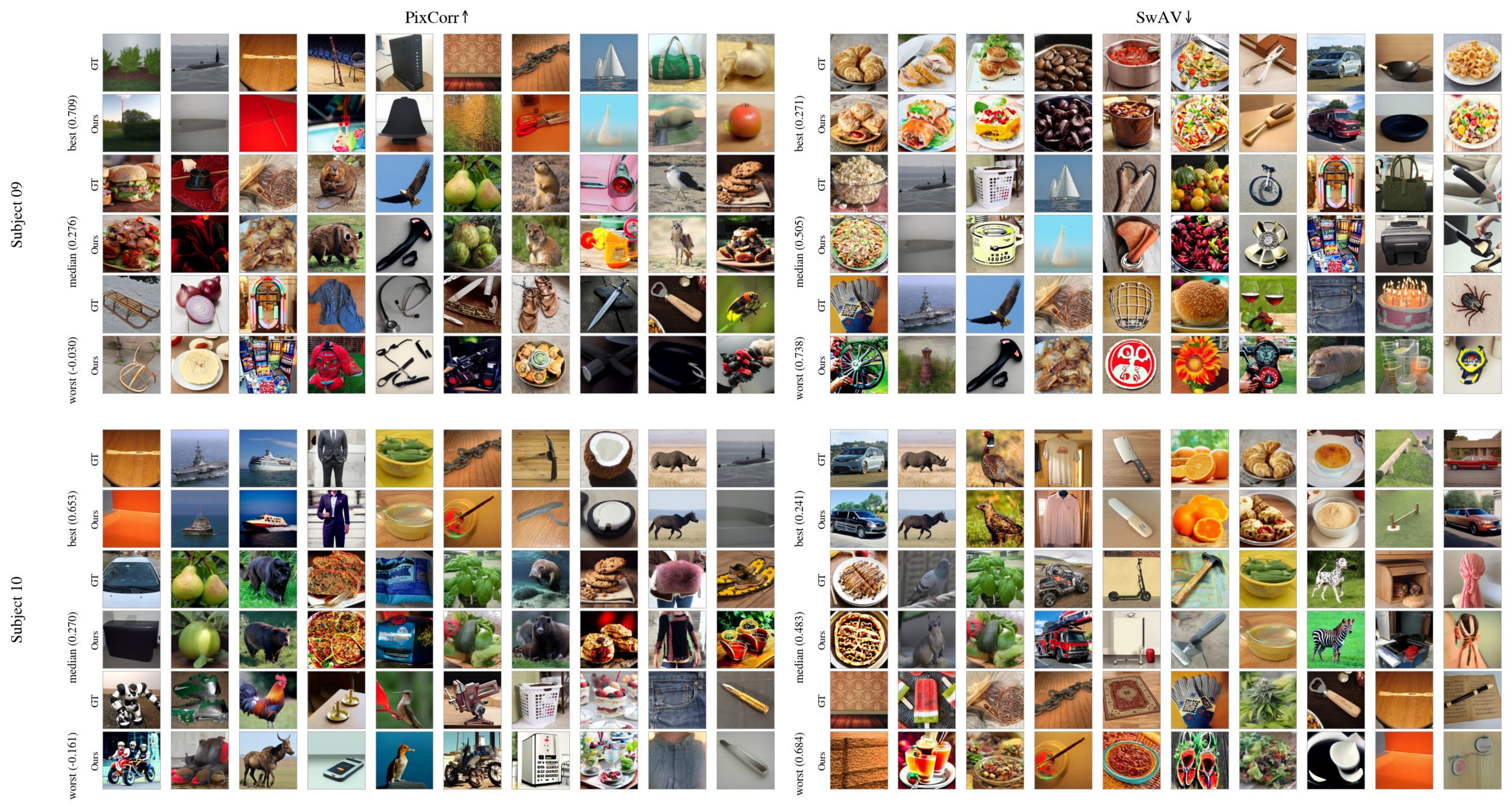}
    \caption{Additional qualitative results for subjects 9-10. For both PixCorr and SwAV, we show the best, median, and worst reconstruction samples selected
    independently from the test set.}
    \label{fig:qr1}
\end{figure*}